\documentclass{cambrian}

\usepackage[numbers,compress]{natbib}

\usepackage{wrapfig}
\usepackage{lineno}
\usepackage{subcaption}
\usepackage{listings}

\usepackage{amsmath,amsfonts,bm}

\def\eqref#1{equation~\ref{#1}}

\def\1{\bm{1}}

\def\vh{{\bm{h}}}

\def\vv{{\bm{v}}}

\def\vx{{\bm{x}}}

\def\vz{{\bm{z}}}

\def\mR{{\bm{R}}}

\DeclareMathAlphabet{\mathsfit}{\encodingdefault}{\sfdefault}{m}{sl}
\SetMathAlphabet{\mathsfit}{bold}{\encodingdefault}{\sfdefault}{bx}{n}

\definecolor{scholarblue}{rgb}{0.21,0.49,0.74}
\definecolor{darkblue}{rgb}{0.83, 0.89, 0.97}
\definecolor{cornellred}{rgb}{0.7, 0.11, 0.11}
\definecolor{cadmiumgreen}{rgb}{0.0, 0.42, 0.24}
\definecolor{aliceblue}{rgb}{0.91, 0.94, 0.97}
\definecolor{Red7}{rgb}{0.941, 0.243, 0.243}
\definecolor{Green7}{RGB}{55, 178, 77}
\definecolor{Blue9}{rgb}{0.098,0.3,0.9}
\definecolor{alphaColor}{HTML}{FF9100}
\definecolor{betaColor}{HTML}{00D5FF}
\definecolor{gammaColor}{HTML}{F0539B}

\hypersetup{
  pagebackref = true,
  breaklinks = true,
  linkcolor = cornellred,
  citecolor  = scholarblue,
  colorlinks = true,
  urlcolor = Blue9
}

\usepackage{amsmath}
\usepackage{amssymb}
\usepackage{booktabs}
\usepackage{multirow}
\usepackage{makecell}
\usepackage{colortbl}
\usepackage{pifont}
\usepackage{graphicx}
\graphicspath{{figures/}} 

\usepackage{pifont}

\usepackage{bm}
\usepackage{algorithm}
\usepackage{algorithmic}

\usepackage{soul}

\usepackage{enumitem}

\usepackage{xspace}

\definecolor{Orchid}{RGB}{218,112,214} %
\definecolor{purple}{RGB}{230,70,151}
\definecolor{lightgray}{gray}{0.9} %

\definecolor{bad}{RGB}{220, 20, 60} 
\definecolor{good}{RGB}{34, 139, 34}

\usepackage{adjustbox}

\usepackage{caption}
\usepackage{subcaption}
\usepackage{graphicx}
\usepackage[dvipsnames]{xcolor} 
\definecolor{nicebg}{rgb}{0.98,0.98,0.98}
\usepackage{bbm}

\usepackage{tikz}
\usetikzlibrary{shapes,arrows,positioning,fit,backgrounds,shadows}
\usepackage{fontawesome5}
\usepackage{tablefootnote}

\usepackage{amsmath}
\usepackage{amssymb}
\usepackage{booktabs}
\usepackage{multirow}
\usepackage{makecell}
\usepackage[table]{xcolor}
\usepackage{graphicx}
\graphicspath{{assets/}} 

\usepackage{pifont}
\newcommand{\cmark}{\ding{51}}
\newcommand{\xmark}{\ding{55}}

\usepackage{enumitem}
\usepackage{bm}
\usepackage{algorithm}
\usepackage{algorithmic}
\usepackage{wrapfig}

\usepackage{soul}

\newcommand{\sref}[1]{\S\ref{#1}}
\newcommand{\fref}[1]{Fig.~\ref{#1}}
\newcommand{\tref}[1]{Tab.~\ref{#1}}

\newcommand{\findingstylename}{finding}
\newcommand{\findingstyle}[1]{\renewcommand{\findingstylename}{#1}}

\newcommand{\findingstylefinding}[3]{%
    \begin{tcolorbox}[
        colback=white!90!gray,
        colframe=teal!60!black,
        arc=5pt,
        boxsep=5pt,
        left=10pt, right=10pt, top=2pt, bottom=2pt,
        boxrule=0.8pt,
        drop shadow=gray!50!white,
        enhanced jigsaw,
        before skip=8pt, after skip=8pt
    ]
    \noindent\faBookmark\hspace{0.5em}\textbf{Finding #1: #2} #3
    \end{tcolorbox}
}

\newcommand{\findingstyletakeaway}[3]{%
    \begin{tcolorbox}[
        colback=blue!3!white,
        colframe=blue!40!black,
        fonttitle=\bfseries,
        title={\faLightbulb\hspace{0.3em}Finding #1: #2},
        arc=2mm,
        boxrule=0.5pt,
        left=5pt, right=5pt, top=2pt, bottom=2pt,
        before skip=10pt, after skip=10pt
    ]
    #3
    \end{tcolorbox}
}

\newcommand{\finding}[3]{%
    \ifx\findingstylename\findingstylefindingname\findingstylefinding{#1}{#2}{#3}%
    \else\findingstyletakeaway{#1}{#2}{#3}\fi
}
\newcommand{\findingstylefindingname}{finding}

\newcommand{\vI}{{\bm{I}}}

\definecolor{edgeblue}{RGB}{0, 0, 200}
\definecolor{edgegreen}{RGB}{0, 200, 0}
\definecolor{gptgreen}{RGB}{0, 166, 126}
\definecolor{scholarpurple}{RGB}{169, 1, 251}
\definecolor{bgcode}{rgb}{0.95,0.95,0.95}
\definecolor{githubgreen}{rgb}{0.564, 0.933, 0.564}
\definecolor{orange}{rgb}{1,0.5,0}

\definecolor{codegreen}{rgb}{0,0.6,0}
\definecolor{codegray}{rgb}{0.5,0.5,0.5}
\definecolor{backcolour}{RGB}{245,248,250}
\definecolor{emph}{RGB}{166,88,53}
\definecolor{nightblue}{RGB}{9,49,105}
\definecolor{keywords}{RGB}{207,33,46}
\definecolor{lightpurple}{RGB}{130,81,223}
\definecolor{examplebg}{RGB}{250,243,240}
\definecolor{codemph}{RGB}{150,30,30}

\newcommand{\epfid}{\ensuremath{\mathrm{EP}_{\mathrm{FID@2}}}}
\newcommand{\epfidk}{\ensuremath{\mathrm{EP}_{\mathrm{FID@}k}}}
\newcommand{\Paragraph}[1]{\par\vspace{2pt}\noindent\textbf{#1}}

\newcommand{\styledquote}[3][scholarblue!8]{%
\begin{center}
\colorbox{#1}{%
    \hspace{0.1in}%
    \begin{minipage}{0.8\textwidth}
    \vspace{0.1in}
    \small\itshape
    #2
    \def\temp{#3}%
    \ifx\temp\empty
    \else
        \begin{flushright}
        \small\normalfont
        ---#3
        \end{flushright}
    \fi
    \vspace{0.1in}
    \end{minipage}%
    \hspace{0.1in}%
}
\end{center}
}

\makeatletter
\newcommand{\smallsym}[2]{#1{\mathpalette\make@small@sym{#2}}}
\newcommand{\make@small@sym}[2]{%
  \vcenter{\hbox{$\m@th\downgrade@style#1#2$}}%
}
\newcommand{\downgrade@style}[1]{%
  \ifx#1\displaystyle\scriptstyle\else
    \ifx#1\textstyle\scriptstyle\else
      \scriptscriptstyle
  \fi\fi
}
\makeatother

\usepackage{xcolor}

\definecolor{mscolor}{rgb}{0.1,0.1,0.9}

\usepackage{xspace} 
\newcommand{\onedot}{.\xspace}

\newcommand{\eg}{\emph{e.g}\onedot}

\usepackage{tcolorbox}
\usepackage{fontawesome5}
\tcbuselibrary{skins,breakable}

\newtcolorbox{takeaway}[1]{
    colback=blue!3!white,
    colframe=blue!40!black,
    fonttitle=\bfseries,
    title={\faLightbulb\hspace{0.3em}Takeaway #1},
    arc=2mm,
    boxrule=0.5pt,
    left=5pt,
    right=5pt,
    top=2pt,
    bottom=2pt,
    before skip=10pt,
    after skip=10pt
}

\newtcolorbox{openquestion}[1]{
    colback=orange!3!white,
    colframe=orange!60!black,
    fonttitle=\bfseries,
    title={\faQuestionCircle\hspace{0.3em}Open Research Question #1},
    arc=2mm,
    boxrule=0.5pt,
    left=5pt,
    right=5pt,
    top=2pt,
    bottom=2pt,
    before skip=10pt,
    after skip=10pt
}

\findingstyle{finding}

\newcommand{\suggestion}[1]{%
    \begin{tcolorbox}[
        colback=white!90!gray,
        colframe=teal!60!black,
        arc=5pt,
        boxsep=5pt,
        left=10pt, right=10pt, top=2pt, bottom=2pt,
        boxrule=0.8pt,
        drop shadow=gray!50!white,
        enhanced jigsaw,
        before skip=8pt, after skip=8pt
    ]
    \noindent\faBookmark\hspace{0.5em}\textbf{Suggestion.} #1
    \end{tcolorbox}
}

\makeatletter
\renewcommand{\abscontent}{
    \noindent
    \centerline{\fontsize{14pt}{14pt}\selectfont\textbf{Abstract}}\vspace{1.5ex}
    \parbox{\dimexpr\linewidth}{\absfont \theabstract}
    \@ifundefined{@keywords}{}{\vskip1em \noindent \keywordsfont Keywords: \@keywords}
}
\renewcommand{\maketitle}{\bgroup\setlength{\parindent}{0pt}
    \vspace*{-6pt}
    \begin{adjustwidth}{0pt}{0pt}
        \begin{flushleft}
            {{\raggedright \titlefont \@title\par}%
             \vskip10pt
             {\raggedright \@author\par}
             \vskip10pt}%
        \end{flushleft}
    \end{adjustwidth}
    \egroup
    {{\abscontent}}%
    \thispagestyle{firststyle}
}
\makeatother

\title{\center{Improved Baselines with Representation Autoencoders}}

\renewcommand\Affilfont{\normalfont\fontsize{11}{15}\selectfont\centering}
\author{%
    \parbox{\textwidth}{%
        \begin{center}
            Jaskirat~Singh\textsuperscript{1,2} \quad\
            Boyang~Zheng\textsuperscript{3} \quad\
            Zongze~Wu\textsuperscript{1} \quad\
            Richard~Zhang\textsuperscript{1} \\
            \vspace{-5pt}
            Eli~Shechtman\textsuperscript{1} \quad\
            Saining~Xie\textsuperscript{3} \\
            \vspace{6pt}
            \textsuperscript{1}Adobe Research \quad \textsuperscript{2}ANU \quad \textsuperscript{3}New York University
        \end{center}
    }%
}

\begin{document}

\fancypagestyle{firststyle}{%
    \fancyhead[L]{}\fancyhead[C]{}\fancyhead[R]{}%
    \fancyfoot[L]{\footerfont\textbf{Project Page:}~\href{https://raev2.github.io}{https://raev2.github.io}}%
    \fancyfoot[C]{}\fancyfoot[R]{}%
    \renewcommand{\headrulewidth}{1pt}%
    \renewcommand{\footrulewidth}{1pt}%
}

\begin{abstract}

Representation Autoencoders (RAE) replace traditional VAE with pretrained vision encoders. 
In this paper, we systematically investigate several design choices and find three insights which simplify and improve RAE. First, we study a generalized formulation where the representation is defined as sum of the last $k$ encoder layers rather than solely the final layer. This simple change 
greatly improves reconstruction without encoder finetuning or specialized data (e.g., text, faces).
Second, we study the prevalent assumption that RAE (using pretrained representation as encoder) replaces representation alignment (REPA), which distills \emph{same representation} to intermediate layers instead. Through large-scale empirical analysis, we uncover a surprising finding: RAE and REPA exhibit \textit{complementary} working mechanisms, allowing same representation to be used as both encoder and target for intermediate diffusion layers. 
Finally, the original RAE struggles with classifier-free guidance (CFG) and requires training a second, weaker diffusion model for AutoGuidance (AG). We show that REPA itself can be viewed as x-prediction in RAE latent space. 
By simply re-parameterizing the output of DiT model, it can provide guidance for ``free''.
Overall, RAEv2 leads to more than $10\times$ faster convergence over original RAE, achieving a state-of-the-art gFID of $1.06$ in just 80 epochs on ImageNet-256. On FDr$^k$ RAEv2 achieves state-of-art 2.17 at just 80 epochs compared to previous best 3.26 (800 epochs) without any post-training. This motivates $\epfidk$ (epochs to reach unguided gFID $\le k$) as a measure of training efficiency. RAEv2 attains an $\epfid$ of 35 epochs, versus 177 for the original RAE.
We also validate our approach across diverse settings for text-to-image generation and navigation world models, showing consistent improvements.
We hope that this work provides useful insights for 
practical adoption of representation autoencoders.

\end{abstract}

\maketitle

\fancyhead[C]{\footerfont Improved Baselines with Representation Autoencoders}

\begin{figure*}[t]
\centering
\includegraphics[width=\textwidth]{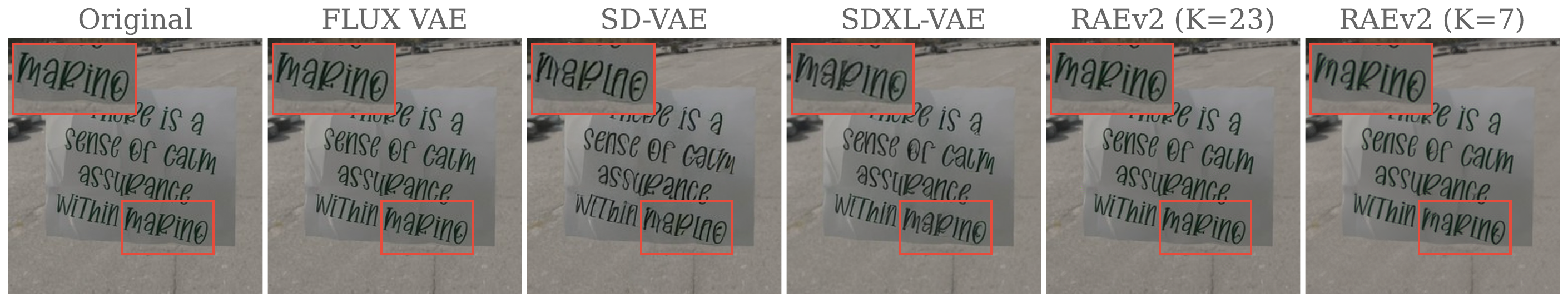}
\vskip -0.1in
\includegraphics[width=\textwidth]{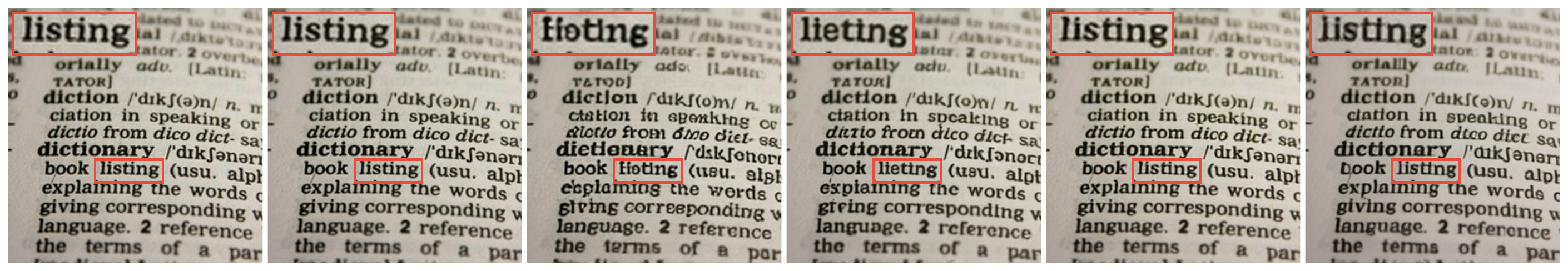}
\vskip -0.05in
\caption{\textbf{Qualitative reconstruction comparisons} with additional data for decoder training \cite{raet2i} (pretrained vision encoder remains frozen). RAEv2 performs competitively with proprietary VAEs. Results use DINOv3-L as encoder for RAEv2. Please see \tref{tab:rec_comparison} for further quantitative results.}
\label{fig:recon_qualitative_additional_data}
\vskip -0.1in
\end{figure*}

\vspace{0.8em}
\noindent\begin{minipage}{\textwidth}
    \begin{center}
\vspace{-0.2cm}
\begin{minipage}{0.495\textwidth}
    \centering
    \includegraphics[width=\textwidth]{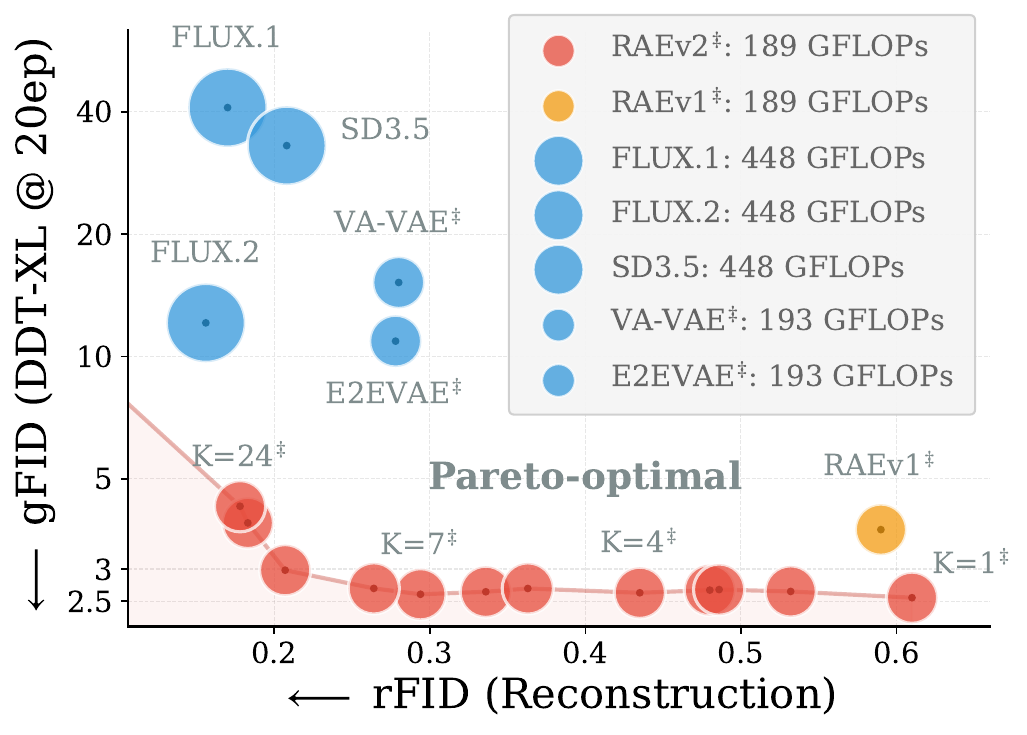}
\end{minipage}
\hfill
\begin{minipage}{0.495\textwidth}
    \centering
    \includegraphics[width=\textwidth]{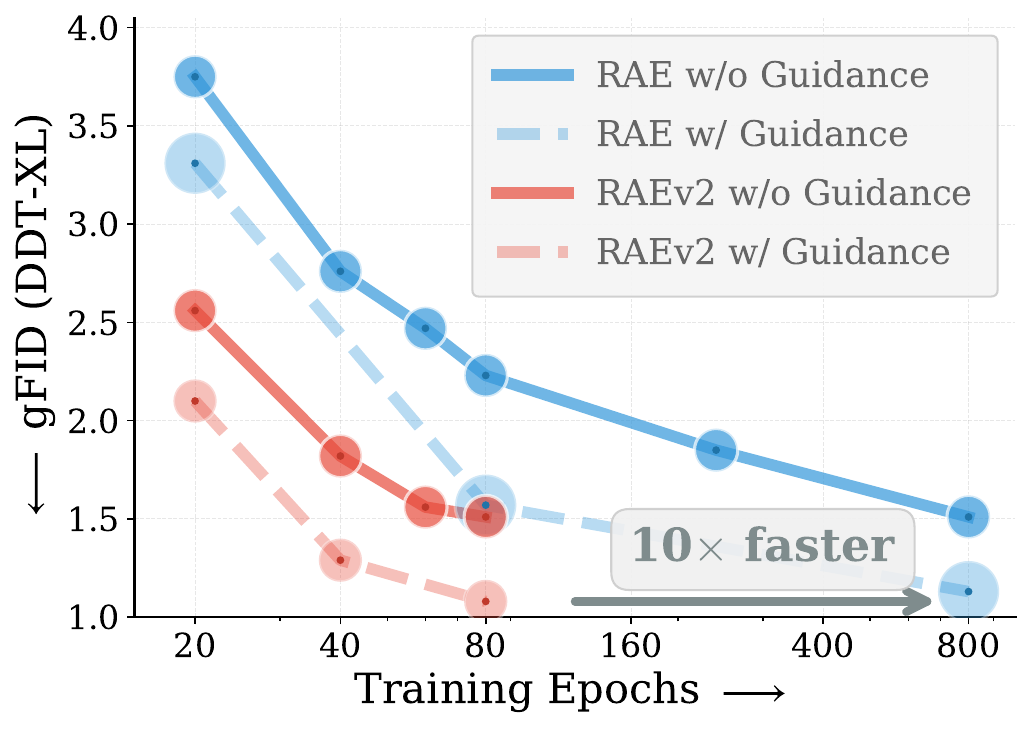}
\end{minipage}
\vskip -0.05in
\captionof{figure}{\textbf{Improved Representation Autoencoders.} \textbf{Left:} RAEv2 exhibits pareto-optimal reconstruction-generation performance at half the encoder FLOPs. $\ddagger$ denotes VAE / RAE / RAEv2 trained only on ImageNet. Training on more data (e.g., text) can further help reconstruction \cite{raet2i} (see \fref{fig:recon_qualitative_additional_data}). \textbf{Right:} Over $10\times$ faster convergence, achieving state-of-the-art gFID of 1.06 in just 80 epochs. }
\label{fig:hero}
\vspace{-0.3cm}
\end{center}

\end{minipage}

\newpage
{
    \hypersetup{linkcolor=black}
    \tableofcontents
}
\newpage

\section{Introduction}
\label{sec:intro}

\begin{figure*}[t]
\centering
\begin{subfigure}[b]{0.24\textwidth}
    \centering
    \includegraphics[width=\textwidth]{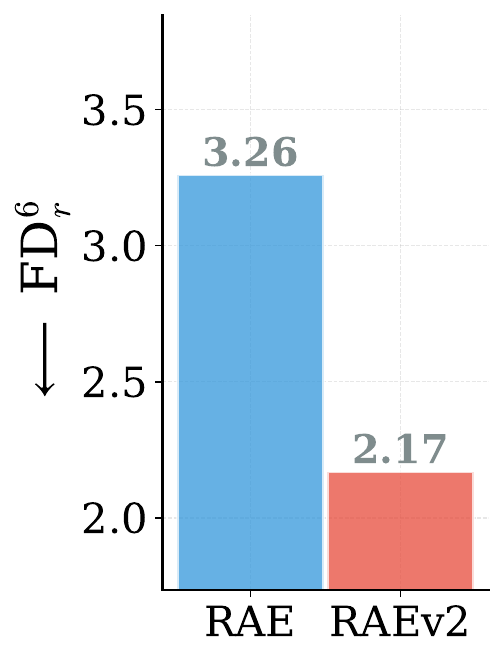}
    \caption{Better generation}
\end{subfigure}
\hfill
\begin{subfigure}[b]{0.24\textwidth}
    \centering
    \includegraphics[width=\textwidth]{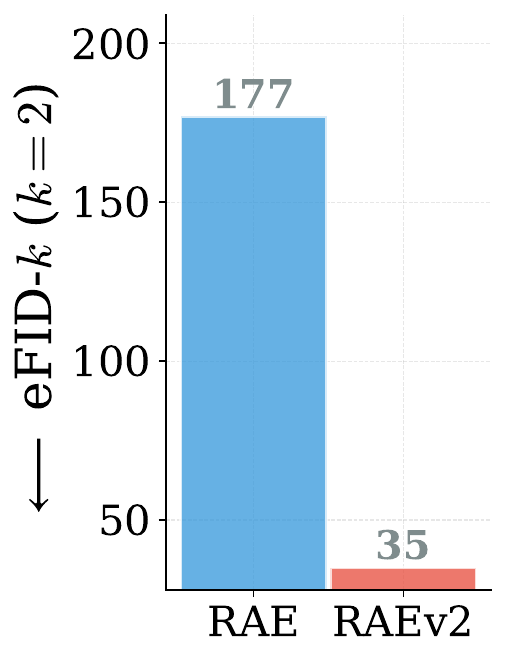}
    \caption{Faster convergence}
\end{subfigure}
\hfill
\begin{subfigure}[b]{0.24\textwidth}
    \centering
    \includegraphics[width=\textwidth]{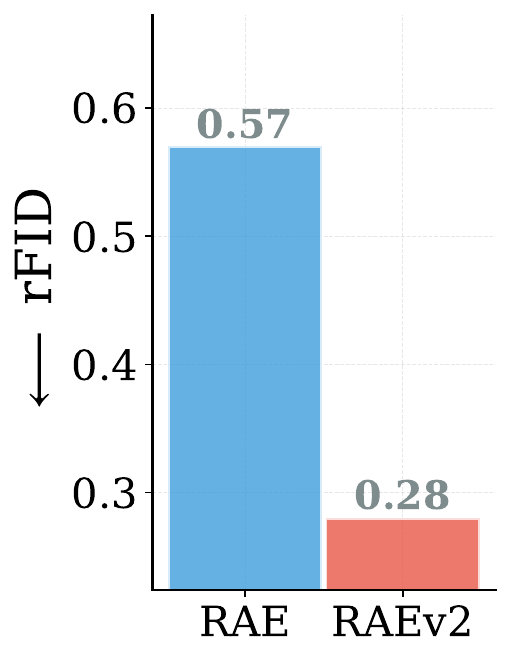}
    \caption{Better reconstruction}
\end{subfigure}
\hfill
\begin{subfigure}[b]{0.24\textwidth}
    \centering
    \includegraphics[width=\textwidth]{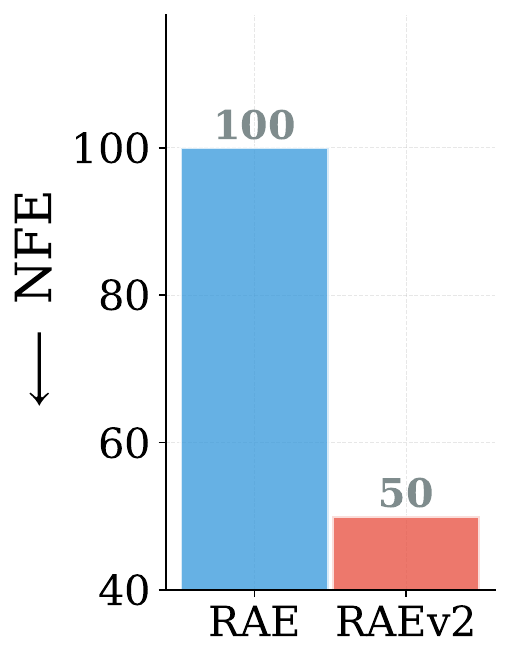}
    \caption{Efficient inference}
\end{subfigure}
\caption{\textbf{Improved performance.} RAEv2 improves over RAE on (a) generation performance: achieving FDr$^6$ \cite{fdr} of 2.17 in just 80 epochs over RAE 3.26 (800 epochs) without any post-training. (b) faster convergence: improving $\epfid$ (epochs to reach unguided gFID $\le 2$) from 177 to 35 (see \sref{subsec:convergence}, \tref{tab:fd_eval}). (c) better reconstruction (d) efficient inference: reusing the REPA head for guidance, eliminating need for separate model (AutoGuidance) and extra forward pass (CFG).}
\label{fig:highlights}
\end{figure*}



Representation Autoencoders (RAE) \cite{rae} have emerged as a powerful framework for replacing traditional VAEs in diffusion transformer training \cite{repa, rae, repae, irepa, fae}, moving a step closer towards a unified tokenization for both understanding and generation. However, several problems persist towards practical adoption: 1) reconstruction performance lags behind specialized VAEs; 2) RAE is incompatible with traditional classifier-free guidance (CFG) \cite{rae}, requiring training a secondary, weaker diffusion model for AutoGuidance \cite{autoguidance}, adding compute and complexity; and 3) the encoder representations themselves remain underexplored, with prior work defaulting to final-layer features.

In this paper, we systematically investigate several design choices and find three key insights which significantly simplify and accelerate RAE training.

\footnotetext{RAEv2 trains within $\sim$10.5 hours on our setup, compared to $>$1 week for 800 epochs in RAE \cite{rae}.}

\begin{figure*}[t]
\centering
\includegraphics[width=\textwidth]{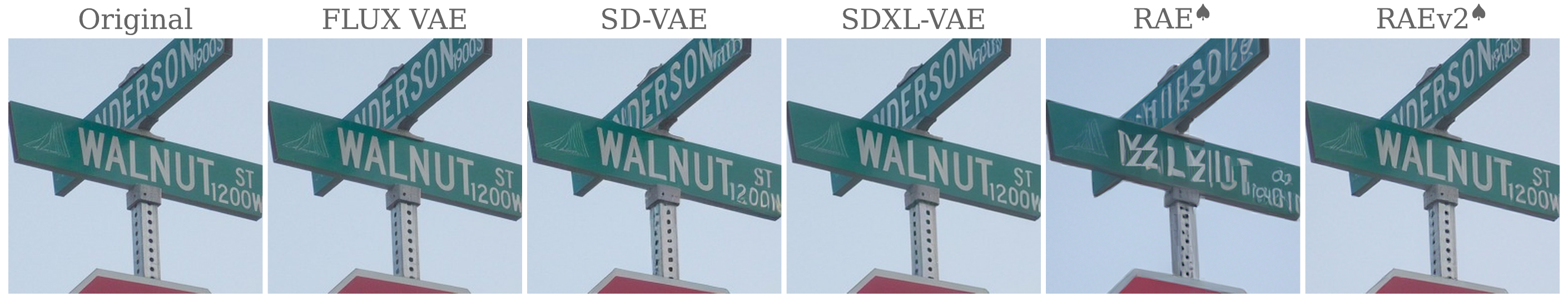}
\vskip -0.1in
\includegraphics[width=\textwidth]{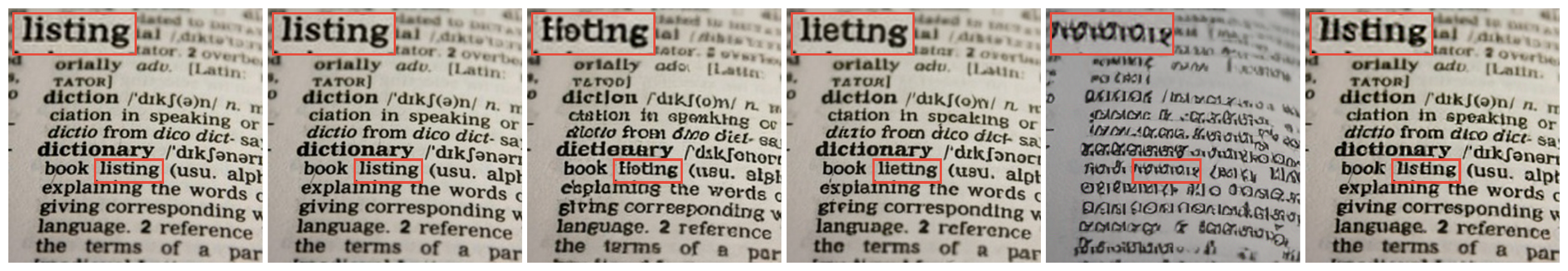}
\vskip -0.05in
\caption{\textbf{Qualitative reconstruction comparison.} $^\spadesuit$ denotes trained only on ImageNet.
RAEv2 despite only being trained on Imagenet performs competitively with proprietary VAEs. Training on more data (e.g., text) can further help reconstruction \cite{raet2i} (see \fref{fig:recon_qualitative_additional_data}). Results use DINOv3-L (K=23) for RAEv2.}

\label{fig:recon_qualitative}
\vskip -0.1in
\end{figure*}

\noindent
\textbf{Generalized Representation Autoencoder.} Prior works typically consider only the final layer output of a pretrained vision encoder as the representation for RAE. However, the representation from a pretrained encoder is not just its final layer; rich and diverse abstractions exist across all layers. We propose a generalized, training-free formulation that simply defines the encoder output as the sum of its last $k$ layers. We find that simply varying $k$ allows easy control over reconstruction quality, leading to Pareto-optimal performance for both reconstruction and generation (Fig.~\ref{fig:hero}, \ref{fig:highlights}, \ref{fig:recon_qualitative}).

\noindent
\textbf{RAE and REPA exhibit complementary working mechanisms.} We next study the prevailing assumption \cite{rae, riprepa, chang2026dino} that RAE (using pretrained representation as latent space encoder) eliminates the need for REPA \cite{repa}, which distills the \emph{same representation} to intermediate diffusion layers. Since RAE already uses encoder features as input, distilling them again to intermediate layers appears to be a wasteful skip connection. We perform large-scale empirical analysis across 27 vision encoders studying the working mechanism of RAE and REPA. The results are surprising: RAE and REPA operate through complementary mechanisms. RAE provides a more semantically rich latent space, while REPA improves the spatial structure of intermediate diffusion features \cite{irepa}. This encourages using the same representation as both encoder (RAE) and target for intermediate layers (REPA). Furthermore, the complementary  mechanism enables stronger encoders (e.g., DINOv3-L) good in both global and spatial performance \cite{irepa, simeoni2025dinov3} to also exhibit better generation performance (\sref{subsec:rae_repa_orthogonal}).


\noindent
\textbf{REPA is x-prediction in RAE latent space.}
The original RAE struggles with traditional classifier-free guidance (CFG), instead relying on AutoGuidance \cite{autoguidance}, which requires training a secondary weaker diffusion model, adding compute and complexity. We observe a key property: when used with RAE, the REPA prediction head performs x-prediction in the target representation space. By simply reformulating the output head as also x-prediction \cite{jit}, we find that the REPA head itself can be used as the weaker baseline for internal-guidance \cite{internalguidance}. This eliminates the need for a separate model entirely (AG). Also unlike CFG, which requires an additional unconditional forward pass (doubling the number of function evaluations at inference), internal-guidance \cite{internalguidance} with REPA head in x-prediction space is computed within the same forward pass, effectively halving the NFEs.



\noindent
\textbf{Training efficiency.}
We combine these insights into an improved baseline RAEv2, which exhibits over $10\times$ faster convergence over original RAE, achieving state-of-art gFID of 1.06 in just 80 epochs. On recently proposed FDr$^k$ metric \cite{fdr} RAEv2 achieves 2.17 in just 80 epochs as opposed to previous best 3.26 (800 epochs) without any post-training. With improved convergence speed of RAEv2, we believe that incremental improvements in the gFID metric might provide little signal for practical applications. Instead the training efficiency of a given method, provides much more useful signal. Motivated by recent speedrun in language domain \cite{modded_nanogpt_2024}, we therefore report $\epfidk$ (epochs to reach unguided gFID $\le k$) as a measure of training efficiency (\tref{tab:fd_eval}). Notably, RAE marks a huge jump over prior works reducing $\epfid$ from 480 to 177. RAEv2 further boosts the training efficiency achieving $\epfid$ of just 35 epochs. We also validate our approach across diverse settings including text-to-image generation and navigation world models \cite{bar2024nwm} (\sref{subsec:generalization}), showing consistent improvements.




\suggestion{Incremental improvements in absolute gFID values might provide limited signal for practical applications. Inspired by recent speedrun in language domain, we also report \emph{training convergence} using $\epfidk$ (epochs to reach unguided gFID $\le k$) (see Table~\ref{tab:fd_eval}).}



\section{Improved Representation Autoencoders}
\label{sec:method}

We next discuss the improved baseline analyzing three insights for improving and simplifying RAE. First, in \sref{subsec:generalized_rae} we generalize the RAE formulation to treat the encoder representation not as a single final-layer feature but as a signal distributed across all layers. This simple change greatly improves reconstruction without encoder finetuning or specialized data (e.g., text, faces) \cite{raet2i}. Next, in \sref{subsec:rae_repa_orthogonal} we perform large-scale empirical analysis finding that RAE and REPA exhibit complementary working mechanisms. 
As a result, using the same representation as both encoder and intermediate target consistently not only improves generation, but also enables stronger encoders (e.g., DINOv3-L) excelling in both global and spatial performance to exhibit better generation with RAEv2.
We then in \sref{subsec:rae_x_prediction} show that REPA when applied with RAE can be viewed as performing x-prediction \cite{jit} in the target latent space. We therefore propose a simple reformulation, which allows the REPA prediction head itself to be used for guidance.

\subsection{Generalized Representation Encoder}
\label{subsec:generalized_rae}

Prior work on RAE usually consider the encoder output as the final-layer feature of a pretrained vision encoder. However, different layers of a pretrained encoder capture complementary features~\cite{bolya2025PerceptionEncoder}. As shown in \fref{fig:per_layer_props}, feature visualizations and spatial self-similarity patterns vary substantially across depth, with later layers emphasizing global semantics and earlier-to-middle layers retaining finer spatial structure. The final layer alone is therefore not always the most informative signal for generation. A natural question arises: \emph{``instead of just relying on the final layer features, can we leverage features \emph{across all layers} without introducing additional parameters or training cost?''}

\begin{figure*}[t]
\centering
\includegraphics[width=\textwidth]{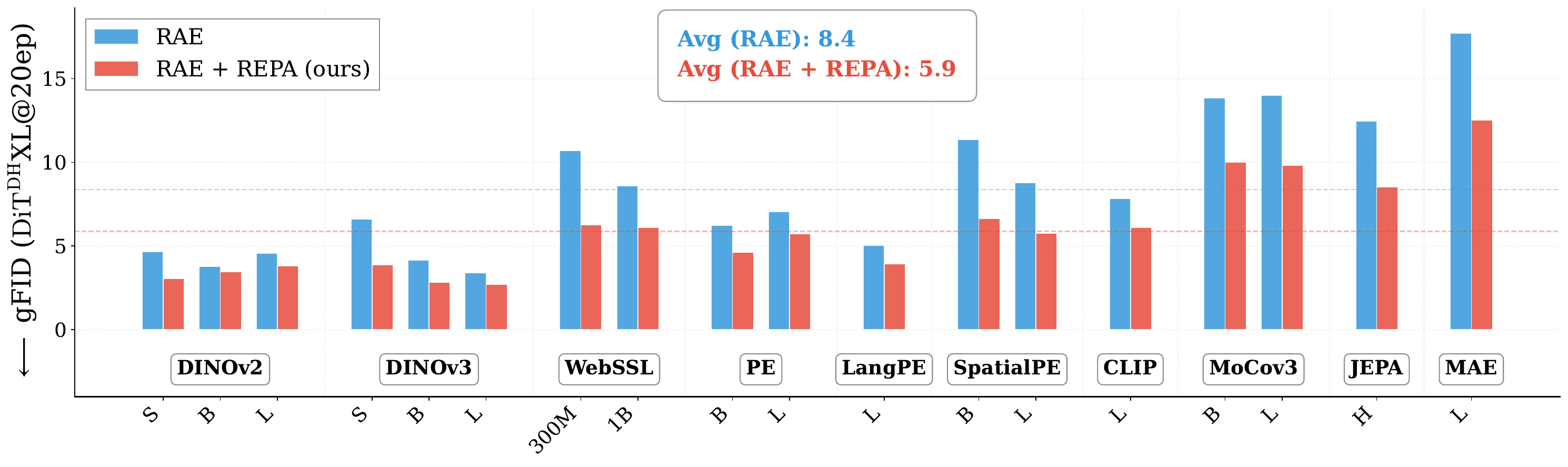}
\vskip -0.05in
\caption{\textbf{RAE does not eliminate need for REPA.} Prevailing assumptions \cite{rae, riprepa, chang2026dino} say that using the pretrained representation (\eg, DINOv2) as both encoder and target of intermediate representations wastes model capacity by introducing a skip connection. Surprisingly, we instead find that RAE and REPA when used together work through complementary working mechanisms (\sref{subsec:rae_repa_orthogonal}). This leads to consistent improvements in generation performance across all pretrained representations. 
}
\label{fig:encoder_sweep}
\vskip -0.1in
\end{figure*}

\Paragraph{Naive concatenation is impractical.} A direct way to use multi-layer features is to concatenate them along the sequence or channel   dimension. For an encoder with $L$ layers producing $N$ tokens of dimension $d$ each, this yields an $LN \times d$ latent sequence. While lossless, this causes an explosion in the latent sequence length, making the resulting latent space substantially more expensive for the diffusion model. On the other hand, concatenation along the channel dimension yields $N \times Ld$ significantly increasing the latent space dimension, making it harder to learn the diffusion model \cite{ldit}.

We instead consider two approaches that combine features across the last $K$ layers while preserving original latent shape $N \times d$. Let $\vz_\ell \in \mathbb{R}^{N \times d}$ denote the feature map at layer $\ell$ of an $L$-layer encoder.
\begin{itemize}[leftmargin=1.5em,itemsep=4pt,topsep=2pt]
    \item \textbf{Simple addition.} The encoder output is defined as the sum of the last $K$ layer features. In high-dimensional spaces, addition preserves the geometric structure of the underlying subspaces~\cite{wiki:dimreduction}:
    \begin{equation}
        \vx \;=\; \sum_{\ell = L-K+1}^{L} \vz_\ell \;\in\; \mathbb{R}^{N \times d}.
    \end{equation}
    \item \textbf{Random-matrix projection.} We concatenate the last $K$ layer features along the channel dimension and project back to $d$ with a fixed random matrix $\mR \in \mathbb{R}^{Kd \times d}$ (sampled once at initialization, \eg i.i.d.\ Gaussian, and held fixed). Random projections are a standard tool in dimensionality reduction~\cite{wiki:dimreduction} and preserve pairwise distances in expectation:
    \begin{equation}
        \vx \;=\; \big[\, \vz_{L-K+1} \,\big\|\, \cdots \,\big\|\, \vz_L \,\big]\, \mR \;\in\; \mathbb{R}^{N \times d}.
    \end{equation}
\end{itemize}

The original RAE is thus a special case in this generalized formulation with $K=1$ i.e., just the final layer. Both approaches keep the latent footprint identical to the original RAE and add no extra learned parameters.  We defer a head-to-head empirical comparison of the two to \sref{sec:experiments}.

\finding{1}{Generalized Representation Encoders.}{Pretrained vision encoders are more than their final layer. Simply aggregating features across layers of a pretrained vision encoder greatly improves reconstruction without encoder finetuning or specialized data 
(e.g., text, faces).
}

\subsection{RAE and REPA exhibit Complementary Working Mechanisms}
\label{subsec:rae_repa_orthogonal}




\Paragraph{Empirical results.} 
We next study the prevailing assumption \cite{rae, riprepa, chang2026dino} that RAE eliminates need for REPA. Since RAE already uses encoder features as input, distilling them again to intermediate layers appears to be a wasteful skip connection. To this end, we first perform large-scale empirical analysis, using the same representation as both encoder and target at intermediate diffusion layers (refer \fref{fig:encoder_sweep}). Results are surprising. Across all encoders, instead of hurting performance, the use of REPA with RAE consistently leads to better generation performance. This suggests a fundamental difference in how representation alignment (REPA) and RAE benefit diffusion training.

\begin{figure*}[t]
\centering
\begin{subfigure}[b]{0.29\textwidth}
\centering
\includegraphics[width=\linewidth]{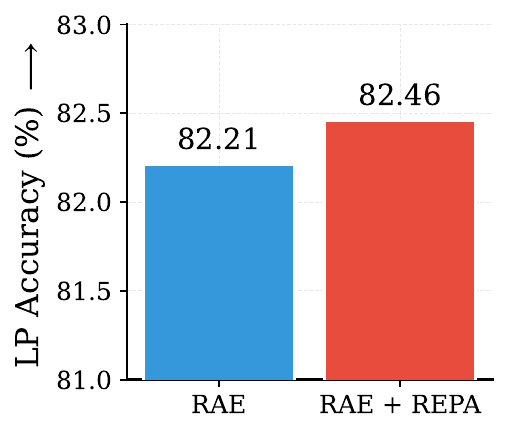}
\caption{Impact of REPA on global semantics with RAE (DINOv2-B)}
\end{subfigure}
\hfill
\begin{subfigure}[b]{0.69\textwidth}
\centering
\includegraphics[width=\linewidth]{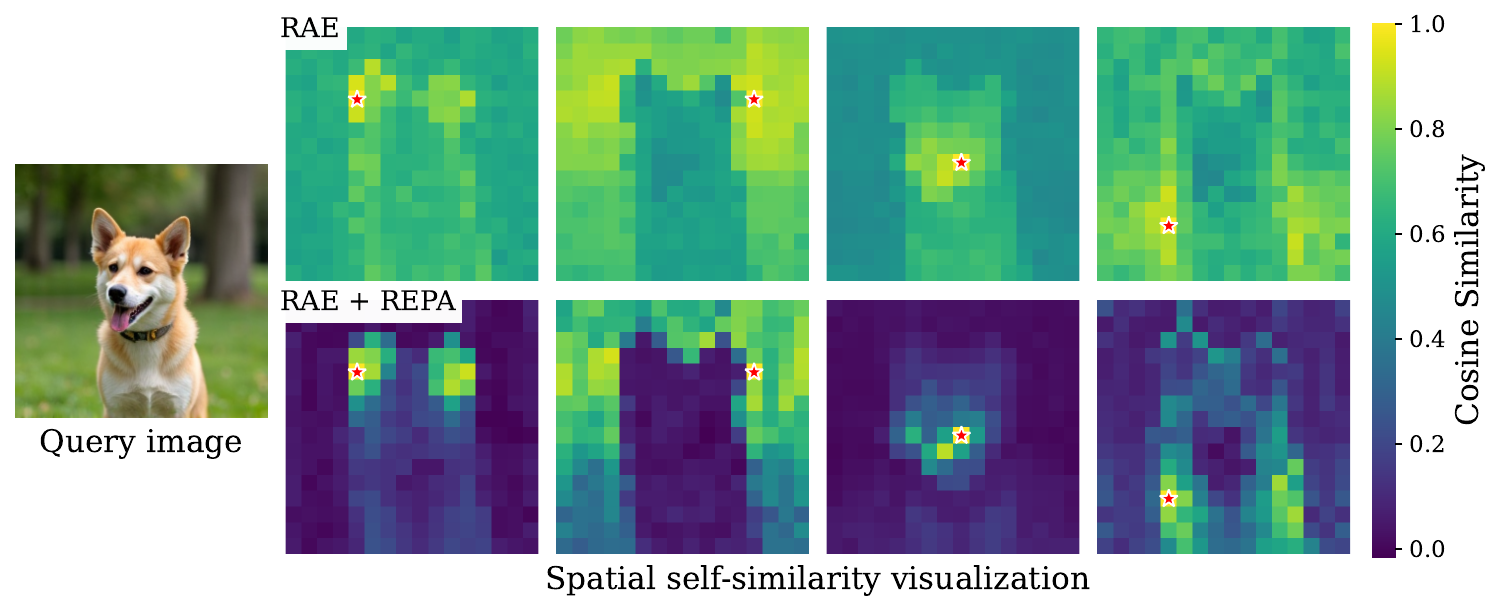}
\caption{Impact of REPA on spatial structure with RAE (DINOv2-B)}
\end{subfigure}
\caption{\textbf{Working mechanism of REPA with RAE.}
While REPA applied with RAE  has minimal impact on global semantics, it significantly improves spatial structure \cite{irepa} of diffusion features.}
\label{fig:working_mechanism}
\vskip -0.2in
\end{figure*}

\Paragraph{Working mechanism.}
We next analyze how REPA impacts diffusion features when combined with RAE.
As shown in \fref{fig:working_mechanism}, adding REPA on top of RAE
has minimal impact on the peak \emph{global semantic information}
(measured through linear probing) of diffusion features.
Instead, we observe that REPA improves the spatial self-similarity
structure of the learned diffusion features (i.e., how different tokens pay attention to each other) - an intriguing phenomenon
recently identified in iREPA~\cite{irepa}.
This suggests complementary working mechanisms for REPA and RAE:
RAE provides a semantically rich latent space for diffusion,
while REPA regularizes the token-token similarity structure
in intermediate diffusion features.

\begin{figure}[h!]
\vskip -0.1in
\centering
\begin{subfigure}[b]{0.49\linewidth}
\centering
\includegraphics[width=\linewidth]{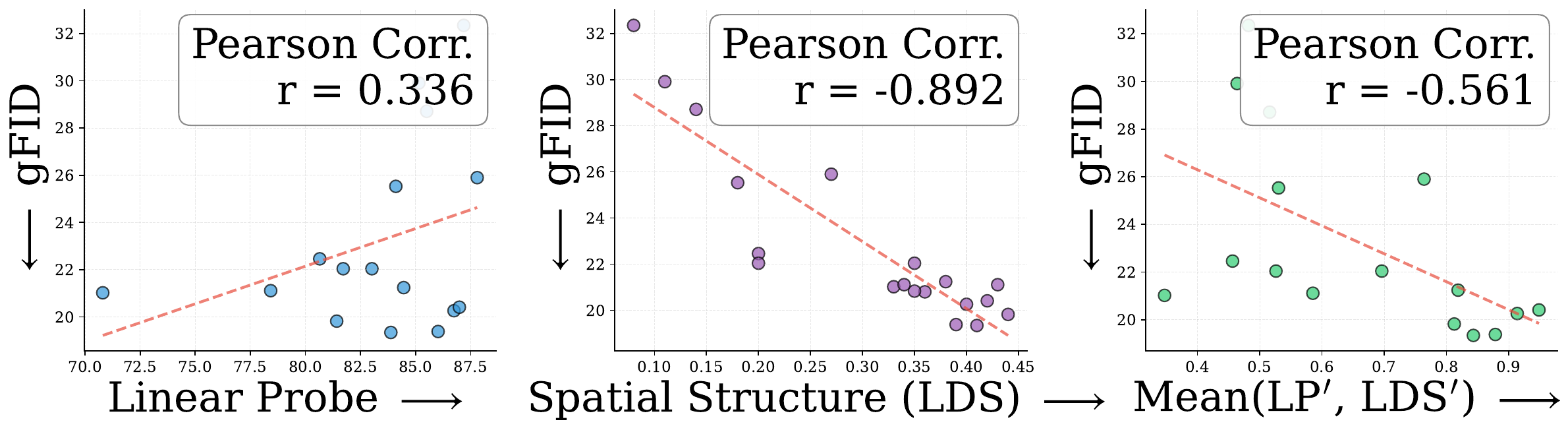}
\caption{REPA alone (SD-VAE)}
\end{subfigure}
\hfill
\begin{subfigure}[b]{0.49\linewidth}
\centering
\includegraphics[width=\linewidth]{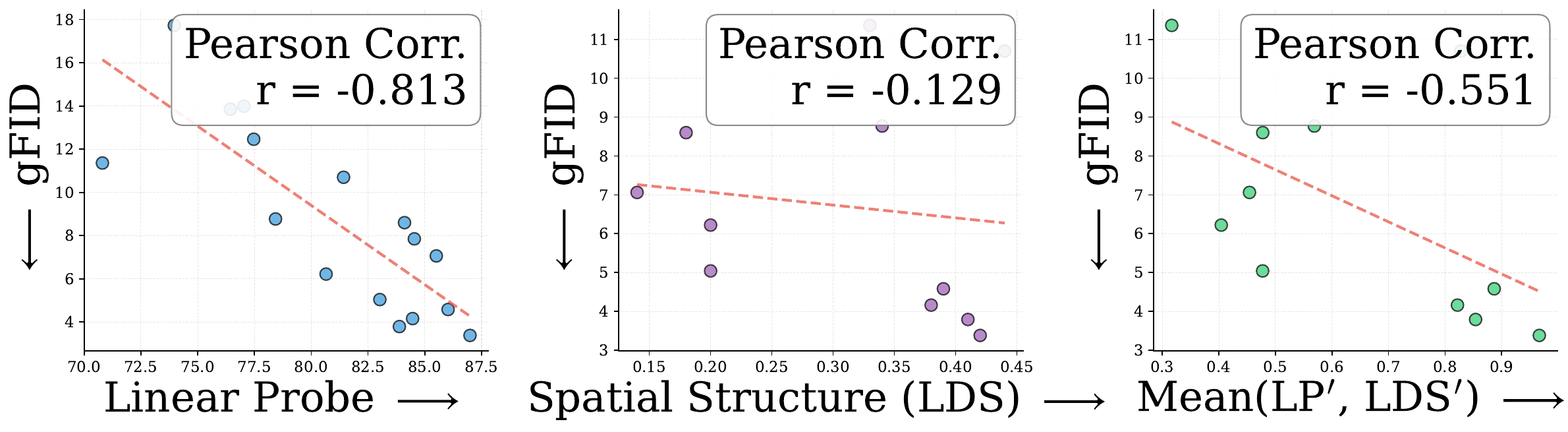}
\caption{RAE alone}
\end{subfigure}

\vskip 0.05in

\begin{subfigure}[b]{0.49\linewidth}
\centering
\includegraphics[width=\linewidth]{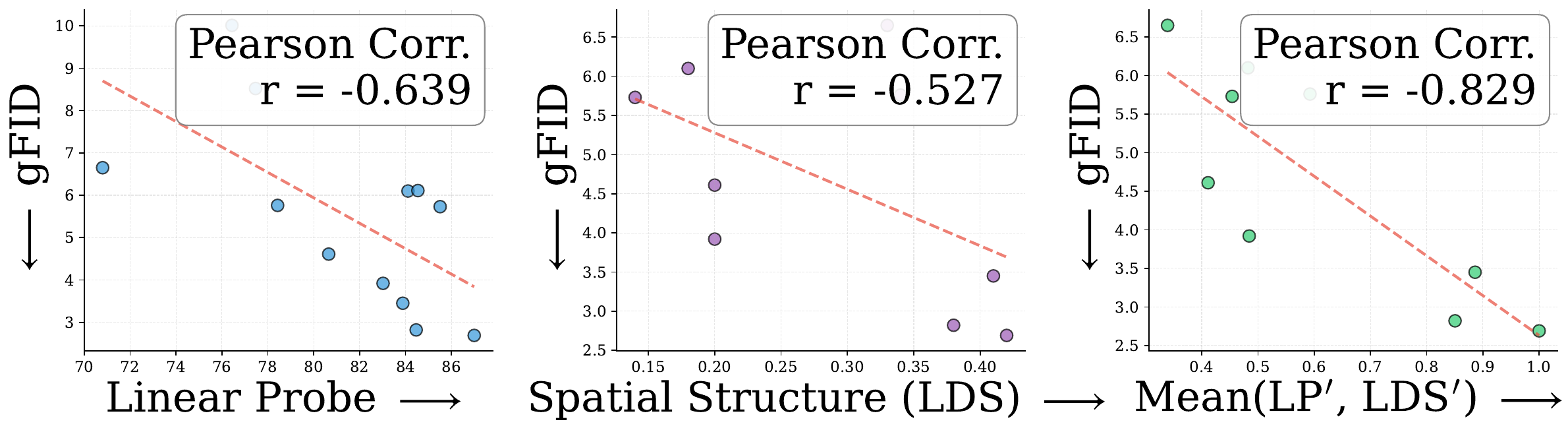}
\caption{RAE + REPA}
\end{subfigure}
\hfill
\begin{subfigure}[b]{0.49\linewidth}
\centering
\scriptsize
\setlength{\tabcolsep}{1.5mm}
\renewcommand{\arraystretch}{1.1}
\begin{tabular}{lccc}
\toprule
Method & LP ($r$)$\downarrow$ & LDS ($r$)$\downarrow$ & Avg ($r$)$\downarrow$ \\
\midrule
REPA alone & +0.34 & \textbf{-0.89} & -0.56 \\
RAE alone  & \textbf{-0.81} & -0.13 & -0.55 \\
\rowcolor{black!5} RAE + REPA & -0.64 & -0.53 & \textbf{-0.83} \\
\bottomrule
\end{tabular}
\caption{Pearson correlation $r$ with gFID}
\end{subfigure}

\caption{\textbf{RAE and REPA leverage complementary encoder properties.} Correlation analysis with gFID across 27 vision encoders for (a) REPA alone, (b) RAE alone, (c) RAE + REPA. (a) Similar to \cite{irepa}, performance with REPA alone correlates more with spatial structure (LDS) \cite{irepa} of a representation. (b) RAE alone benefits much more from higher global semantics (LP). (c) Together, RAE and REPA benefit from encoders strong in both global semantics (LP) and spatial structure (LDS). This explains why stronger encoders (e.g., DINOv3-L) which excel in both global and spatial performance yield the best generation with RAEv2 (Tab.~\ref{tab:encoder}). All results: DDT-XL, 20 epochs, without guidance.}
\label{fig:repr_correlation}
\vskip -0.2in
\end{figure}

\Paragraph{Correlation analysis.}
To further validate the complementary mechanisms of RAE and REPA, we follow the practice in iREPA~\cite{irepa}, analyzing the Imagenet linear probing accuracy (LP) and local distance similarity score (LDS)~\cite{irepa} across 27 vision encoders, and report their Pearson correlation $r$ with generation
quality (gFID). 
As shown in \fref{fig:repr_correlation}, for REPA alone (with VAE),
LDS is highly predictive ($|r|{=}0.89$) while LP is actually
anticorrelated ($r{=}{+}0.34$), consistent with findings
in~\cite{irepa}.
In contrast, when using RAE alone, LP dominates ($|r|{=}0.81$)
while LDS barely correlates ($r{=}|0.13|$).
When combining RAE with REPA, neither metric alone is strongly
predictive, but the average of LP (global semantics) and LDS (spatial structure) achieves the highest
correlation ($|r|{=}0.83$).
This confirms that RAE and REPA operate through complementary
mechanisms: RAE leverages global semantics while REPA regularizes
spatial structure.

\Paragraph{Selecting the best representation.} The above complementarity also enables stronger representations (e.g., DINOv3-L) that perform well for both global (LP) and spatial (LDS) performance, to also exhibit better generation with RAEv2 
We defer the detailed encoder-selection study to \sref{subsec:ablations}.

\finding{2}{RAE and REPA exhibit complementary working mechanisms.}{RAE leverages semantic quality  while REPA regularizes spatial structure. 
This complementary nature allows using same pretrained representation as both encoder (RAE) and target for intermediate diffusion features (REPA). This also explains why stronger representations like DINOv3-L, which excel in both global and spatial performance, achieve the best generation with RAEv2 (see \tref{tab:encoder}).}

\subsection{Reformulating REPA as x-prediction with RAE}
\label{subsec:rae_x_prediction}

We next show that when used with RAE, the REPA head itself can be used for guidance, eliminating need for training second weaker diffusion model (AutoGuidance) or additional forward pass (CFG).


\begin{wraptable}{r}{0.45\textwidth}
\centering
\vskip -0.15in
\scriptsize
\setlength{\tabcolsep}{2.5mm}
\renewcommand{\arraystretch}{1.05}
\begin{tabular}{lcc}
\toprule
Guidance & gFID$\downarrow$ & IS$\uparrow$ \\
\midrule
w/o Guidance              & 3.75 & 198.7 \\
CFG~\cite{cfg}             & 3.86 & 276.4 \\
\rowcolor{black!5} Autoguidance (AG)~\cite{autoguidance} & \textbf{3.31} & 219.1 \\
\bottomrule
\end{tabular}
\vskip -0.05in
\caption{RAE DiT$^{DH}$-XL, 20 epochs.}
\vskip -0.15in
\label{tab:guidance}
\end{wraptable}

\Paragraph{RAE struggles with traditional CFG.}
As shown in \tref{tab:guidance}, we confirm
that original RAE \cite{rae} struggles with standard classifier-free guidance~\cite{cfg}. RAE therefore relies on AutoGuidance~\cite{autoguidance}, a separate, smaller model trained to serve as the ``weaker'' baseline, adding compute and complexity.

\Paragraph{REPA is x-prediction in RAE latent space.}
We observe a key connection. In RAE, the clean latent \emph{is} the encoder representation: $\vx = E(\vI)$. The REPA projection head $h_\phi$ maps early-layer intermediate features $\vh$ to predict the clean latent $\hat{\vx}_{\text{repa}} = h_\phi(\vh)$. This is exactly x-prediction~\cite{jit} in the RAE latent space. Importantly, because $h_\phi$ is a lightweight MLP that only accesses early-layer features, its prediction is inherently weaker than the full model's, playing the same role as the separately trained smaller model in AutoGuidance~\cite{autoguidance}.

\Paragraph{Reformulating REPA head for guidance.}
If we reformulate the full model output to also give x-prediction instead of velocity \cite{dit}, both outputs live in the same space. Let 
$\hat{\vx}_{\text{full}}$
denote the full model's x-prediction (all layers) and 
$\hat{\vx}_{\text{repa}}$ 
the REPA head's x-prediction (early layers only). We can then apply internal-guidance \cite{internalguidance} directly as,
\begin{equation}
\hat{\vx}_{\text{guided}} = \hat{\vx}_{\text{full}} + w \cdot (\hat{\vx}_{\text{full}} - \hat{\vx}_{\text{repa}}),
\label{eq:self_guidance}
\end{equation}
and convert back to velocity for sampling or loss computation: $\vv = (\vx_t - \hat{\vx}_{\text{guided}}) / t$. The REPA head runs during the same forward pass as main model, so this eliminates the need for training a separate weaker model (AG \cite{autoguidance}) and no additional forward pass (CFG \cite{cfg}). 

Thus, \emph{when used with RAE} our formulation is equivalent to a deep supervised network \cite{lee2015deeply} or internal-guidance \cite{internalguidance},
with additional reparameterization to x-prediction. The reparameterization to x-prediction \cite{jit} is important as it allows use of REPA-head for both supervising spatial structure of intermediate layers (\sref{subsec:rae_repa_orthogonal}) and act as a weaker baseline for guidance after reparametrization. Please see \tref{tab:xpred_ablation} for ablation on importance of reparameterization to x-prediction.

\finding{3}{REPA enables self-guidance.}{
REPA is x-prediction in RAE latent space. By reformulating output head also as x-prediction, REPA head itself can be used for internal-guidance; eliminating need for a separate model (AG) or extra forward pass (CFG).}



\section{Experiments}
\label{sec:experiments}

We validate the performance of our approach through extensive
experiments on ImageNet, text-to-image generation and world models. In particular, we investigate the following research questions:
\begin{itemize}[leftmargin=*,itemsep=0mm]
\item Does the improved training recipe consistently improve convergence speed with representation autoencoders across diverse settings, model scales etc? (Fig.~\ref{fig:hero}, \ref{fig:highlights}, \ref{fig:encoder_sweep}, \ref{fig:convergence}, \ref{fig:rfid_gfid}; Tab.~\ref{tab:encoder}, \ref{tab:convergence_guidance}, \ref{tab:model_scale}, \ref{tab:fd_eval})
\item Can we use generalized RAE formulation for improving reconstruction performance of representation autoencoders in a training free manner? (Fig.~\ref{fig:hero}, \ref{fig:recon_qualitative}, \ref{fig:k_sweep}, \ref{fig:rfid_gfid}; Tab.~\ref{tab:genrae_formulation}, \ref{tab:rec_comparison})
\item Does the proposed approach generalize to diverse training settings including text-to-image generation and world models? (Fig.~\ref{fig:qualitative_t2i_main}, \ref{fig:nwm_horizon}, \ref{fig:nwm_qualitative}; Tab.~\ref{tab:nwm_fvd}, \ref{tab:t2i})
\end{itemize}






\subsection{Ablation Studies}
\label{subsec:ablations}

We first ablate different design choices for different components proposed in \sref{sec:method} on ImageNet-256. Unless otherwise specified we use DiT$^{DH}$-XL, DINOv3-L as encoder and batch size 1024.

\noindent
\sethlcolor{green!10}\hl{\textbf{Encoder selection.}} Results are shown in \tref{tab:encoder}. The original RAE~\cite{rae} uses DINOv2-B as its encoder because it gives the best generation among the encoders tested under the RAE recipe. With RAEv2, however, the picture changes: stronger representations such as DINOv3-B~\cite{simeoni2025dinov3} yield better generation, despite performing worse than DINOv2-B under the original RAE recipe. This is consistent with correlation analysis in \sref{subsec:rae_repa_orthogonal}; stronger representations (e.g., DINOv3-L) which excel in both global semantics and spatial performance lead to best generation with RAEv2. Based on this finding, we use DINOv3-L as the default encoder for all subsequent RAEv2 experiments.


\noindent
\sethlcolor{Plum!10}\hl{\textbf{Formulation for generalized RAE.}}
In \tref{tab:genrae_formulation}, we compare the two parameter-free aggregation schemes from \sref{subsec:generalized_rae} on DINOv3-L: simple addition of the last $K$ encoder layers (\textbf{MLS}) versus fixed random-matrix projection of their channel-wise concatenation (\textbf{MLR}). 
Interestingly, while both methods are effectively tied on Stage-1 reconstruction, MLS consistently wins on Stage-2 performance. We therefore use MLS as the default aggregation in the rest of the paper.

\begin{table}[h!]
\centering
\small
\setlength{\tabcolsep}{3mm}
{
\begin{tabular}{lcccccc}
\toprule
\multirow{2}{*}{Encoder} & \multicolumn{3}{c}{Encoder properties} & \multicolumn{2}{c}{gFID (DiT$^{\text{DH}}$-XL @ 20ep) $\downarrow$} \\
\cmidrule(lr){2-4} \cmidrule(lr){5-6}
 & LP $\uparrow$ & LDS $\uparrow$ & Avg(LP', LDS') $\uparrow$ & RAE & RAEv2 ($k{=}1$) \\
\midrule
MoCov3-B~\cite{mocov3}            & 76.4 & 0.15 & 0.46 & 13.84 & 8.35 \\
WebSSL-1B~\cite{fan2025scaling}   & 84.1 & 0.18 & 0.51 & 8.60  & 4.16 \\
DINOv2-B~\cite{dinov2}            & 83.9 & 0.41 & 0.62 & 3.75  & 2.81 \\
DINOv3-B~\cite{simeoni2025dinov3} & 84.5 & 0.38 & 0.61 & 4.25  & 2.76 \\
\rowcolor{black!5} DINOv3-L~\cite{simeoni2025dinov3} & \textbf{87.0} & \textbf{0.42} & \textbf{0.65} & \textbf{3.30} & \textbf{2.61} \\
\bottomrule
\end{tabular}
}
\vskip 0.1in
\caption{\sethlcolor{green!10}\hl{\textbf{Ablation on choice of pretrained vision encoder.}} gFID at 20 epochs (DiT$^{\text{DH}}$-XL). We observe that with RAEv2, stronger encoders e.g, DINOv3-L with both better global (LP) and spatial (LDS)~\cite{irepa} representations achieve the best performance. Please refer \tref{tab:encoder_appendix} for further results.}
\label{tab:encoder}
\end{table}

\begin{table}[t]
\centering
\begin{minipage}[t]{0.4\textwidth}
\centering
\small
\setlength{\tabcolsep}{2.5mm}
\renewcommand{\arraystretch}{1.1}
\newcommand{\gb}{\cellcolor{gray!15}}
\begin{tabular}{c l cc}
\toprule
$K$ & Method & rFID $\downarrow$ & gFID $\downarrow$ \\
\midrule
\multirow{2}{*}{2} & MLR      & 0.570       & 3.085 \\
                   & \gb{}MLS & \gb{}0.532  & \gb{}\textbf{2.586} \\
\midrule
\multirow{2}{*}{8} & MLR      & 0.268       & 3.580 \\
                   & \gb{}MLS & \gb{}0.264  & \gb{}\textbf{2.688} \\
\bottomrule
\end{tabular}
\vskip 0.05in
\caption{\sethlcolor{Plum!10}\hl{\textbf{Ablation on Generalized RAE formulation.}} MLS dominates MLR for gfid (see \sref{subsec:ablations}). Full sweep in \tref{tab:genrae_formulation_appendix}.}
\label{tab:genrae_formulation}
\end{minipage}

\hfill
\begin{minipage}[t]{0.55\textwidth}
\centering
\small
\setlength{\tabcolsep}{1.5mm}
\renewcommand{\arraystretch}{1.1}
\begin{tabular}{l cc}
\toprule
Guidance & gFID ($K{=}7$) $\downarrow$ & gFID ($K{=}23$) $\downarrow$ \\
\midrule
w/o Guidance                          & 1.65          & 3.01          \\
CFG~\cite{cfg}                        & 1.49          & 2.83          \\
Autoguidance (AG)~\cite{autoguidance} & 1.14          & 1.37          \\
\rowcolor{black!5}
REPA Guidance                 & \textbf{1.06} & \textbf{1.25} \\
\bottomrule
\end{tabular}
\vskip 0.05in
\caption{\sethlcolor{yellow!10}\hl{\textbf{Ablation on Guidance mechanism in RAEv2.}} Guidance with REPA and x-prediction achieves best results at no extra inference cost. Full results in \tref{tab:convergence_guidance_appendix}.}
\label{tab:convergence_guidance}
\end{minipage}

\end{table}

\begin{figure}[t]
\centering
\begin{subfigure}[b]{0.24\linewidth}
\includegraphics[width=\linewidth]{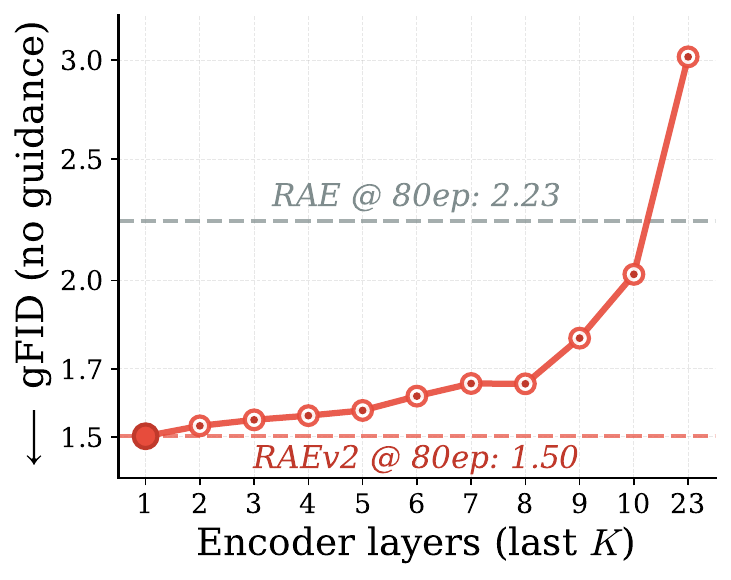}
\subcaption{gFID $\downarrow$ (no guidance)}
\label{fig:k_sweep_gfid_noguide}
\end{subfigure}
\hfill
\begin{subfigure}[b]{0.24\linewidth}
\includegraphics[width=\linewidth]{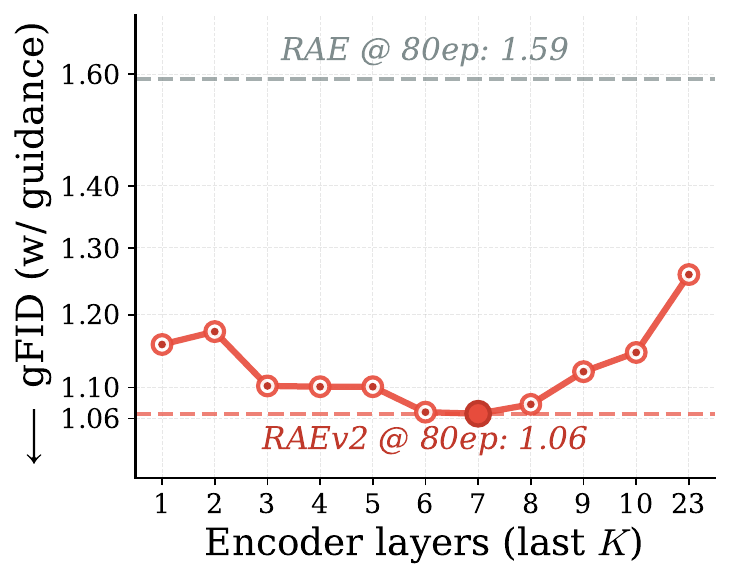}
\subcaption{gFID $\downarrow$ (with guidance)}
\label{fig:k_sweep_gfid_guide}
\end{subfigure}
\hfill
\begin{subfigure}[b]{0.24\linewidth}
\includegraphics[width=\linewidth]{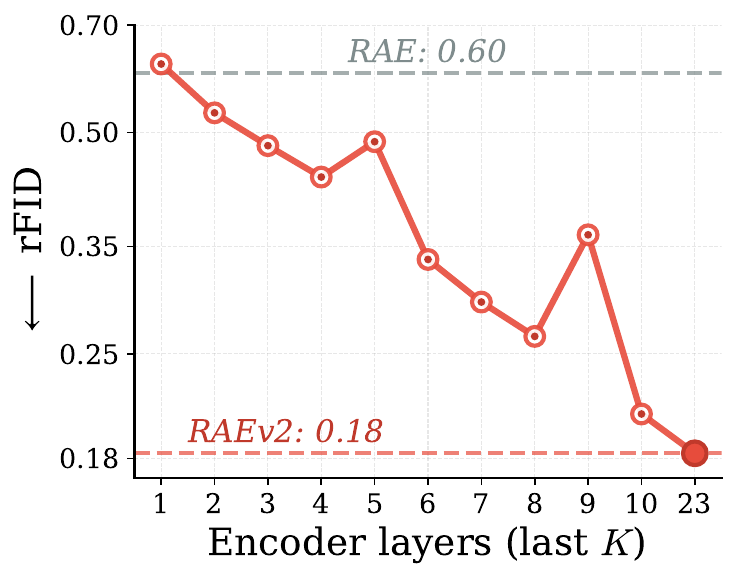}
\subcaption{rFID $\downarrow$}
\label{fig:k_sweep_rfid}
\end{subfigure}
\hfill
\begin{subfigure}[b]{0.24\linewidth}
\includegraphics[width=\linewidth]{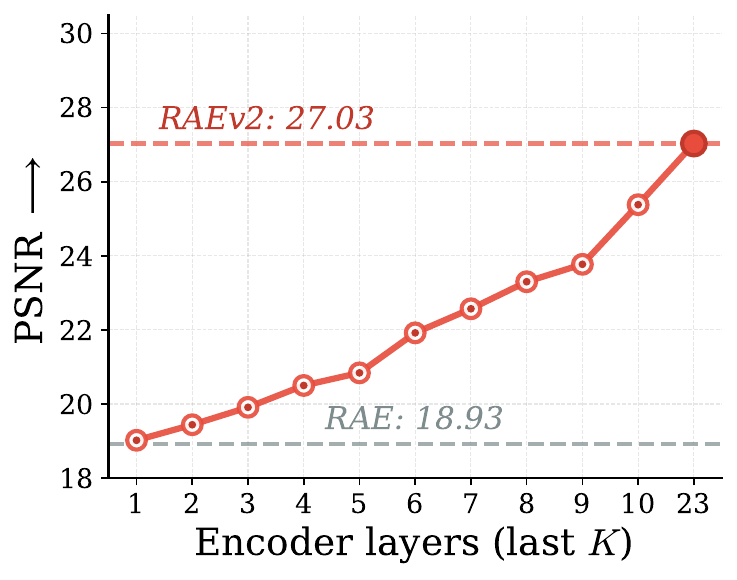}
\subcaption{PSNR $\uparrow$}
\label{fig:k_sweep_psnr}
\end{subfigure}
\caption{\sethlcolor{cyan!10}\hl{\textbf{Ablation on choice of $K$ for generalized RAE.}} (a, b)~Stage-2 generation quality without and with guidance. (c, d)~Stage-1 reconstruction (rFID and PSNR). All results with DINOv3-L (24 layers), DDT-XL and 80 epochs.
Stage-1 reconstruction (rFID, PSNR) improves monotonically with $K$. Interestingly, the generalized RAE not only
improves reconstruction performance but also leads to better generation performance with guidance (best at $K=7$).}
\label{fig:k_sweep}
\vskip -0.15in
\end{figure}

\noindent
\sethlcolor{cyan!10}\hl{\textbf{Choice of $K$ for generalized RAE.}}
We sweep $K \in \{1, \dots, 10, 23\}$ for generalized RAE on DINOv3-L (\fref{fig:k_sweep}). Stage-1 reconstruction (rFID, PSNR) improves monotonically with $K$, rFID 0.18 and PSNR 27.03 at $K{=}23$, well past the standard RAE baseline (rFID 0.60, PSNR 18.93). Stage-2 generation behaves differently: at just 80 epochs, the unguided gFID is best near $K{=}1$ (1.50), while the guided gFID performs best with $K{=}7$ (1.06). Thus, interestingly the generalized RAE not only improves reconstruction performance but also leads to better generation performance with guidance.


\begin{wraptable}{r}{0.35\textwidth}
\centering
\vskip -0.15in
\scriptsize
\setlength{\tabcolsep}{2mm}
\renewcommand{\arraystretch}{1.05}
\begin{tabular}{l c}
\toprule
$K$ & LP top-1 (\%) $\uparrow$ \\
\midrule
1 (last layer; RAE) & 85.39 \\
4                   & 85.15 \\
7                   & 85.10 \\
23 (full MLS)       & 85.24 \\
\bottomrule
\end{tabular}
\vskip -0.05in
\caption{Linear probing on ImageNet across $K$ (DINOv3-L); 30 epochs of LP training, further training may improve scores. Full sweep in \tref{tab:genrae_lp}.}
\label{tab:genrae_lp_main}
\vskip -0.2in
\end{wraptable}
\noindent
\sethlcolor{Lavender!30}\hl{\textbf{Impact of generalized RAE on understanding performance.}}
A key advantage of RAE is that it provides a unified tokenization for both understanding and generation. We study the impact of the generalized formulation on the encoder's understanding performance with different $K$ in \tref{tab:genrae_lp_main} ($K{=}1$ is the original RAE). The generalized formulation improves reconstruction and guided generation (\fref{fig:k_sweep}) while preserving linear probing performance on ImageNet. Full sweep over $K \in \{1, \dots, 10, 23\}$ is in \tref{tab:genrae_lp}.

\finding{4}{}{
Generalized formulation of RAE helps improve reconstruction and generation performance with guidance (\fref{fig:k_sweep}) while preserving global semantics of the representation space (\tref{tab:genrae_lp_main}). This enables its use for a unified tokenization for both understanding and generation.}

\noindent
\sethlcolor{yellow!10}\hl{\textbf{Guidance mechanism in RAEv2.}}
We ablate four guidance options for RAEv2 in \tref{tab:convergence_guidance}: (i) no guidance, (ii) classifier-free guidance (CFG)~\cite{cfg}, (iii) AutoGuidance~\cite{autoguidance}, and (iv) internal guidance~\cite{internalguidance} with REPA-head and x-prediction (\sref{subsec:rae_x_prediction}). CFG fails to meaningfully improve gFID, AG helps but requires an additional model and forward pass. In contrast, internal guidance with REPA-head achieves the best gFID at no extra inference cost.

\subsection{Impact on Convergence Speed}
\label{subsec:convergence}

\begin{figure*}[t]
\centering
\begin{subfigure}[b]{0.32\textwidth}
\centering
\includegraphics[width=\linewidth]{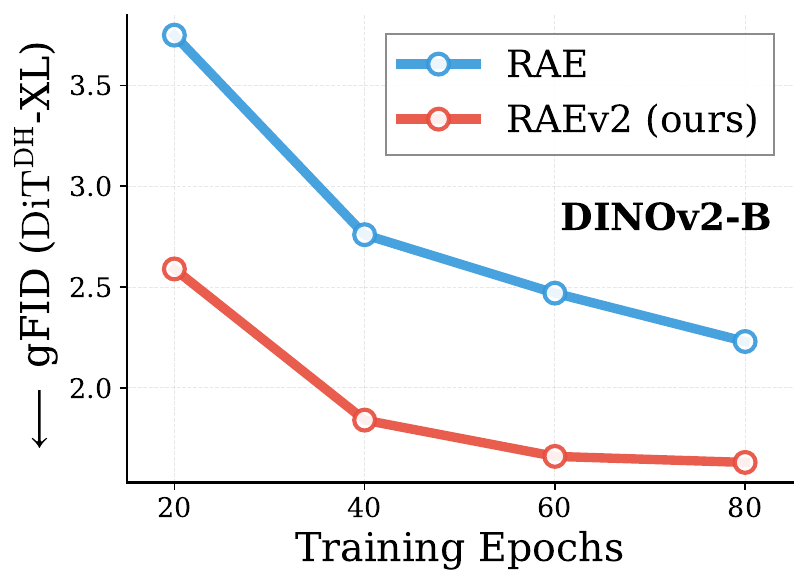}
\end{subfigure}
\hfill
\begin{subfigure}[b]{0.32\textwidth}
\centering
\includegraphics[width=\linewidth]{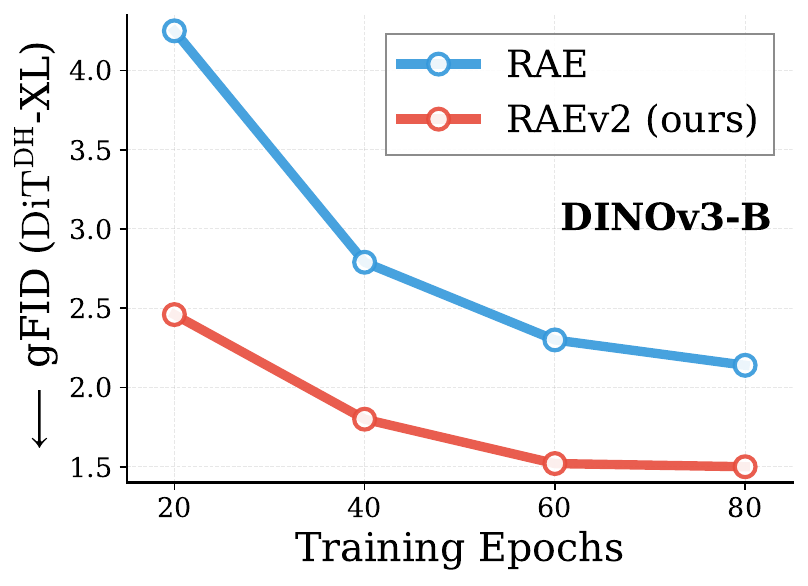}
\end{subfigure}
\hfill
\begin{subfigure}[b]{0.32\textwidth}
\centering
\includegraphics[width=\linewidth]{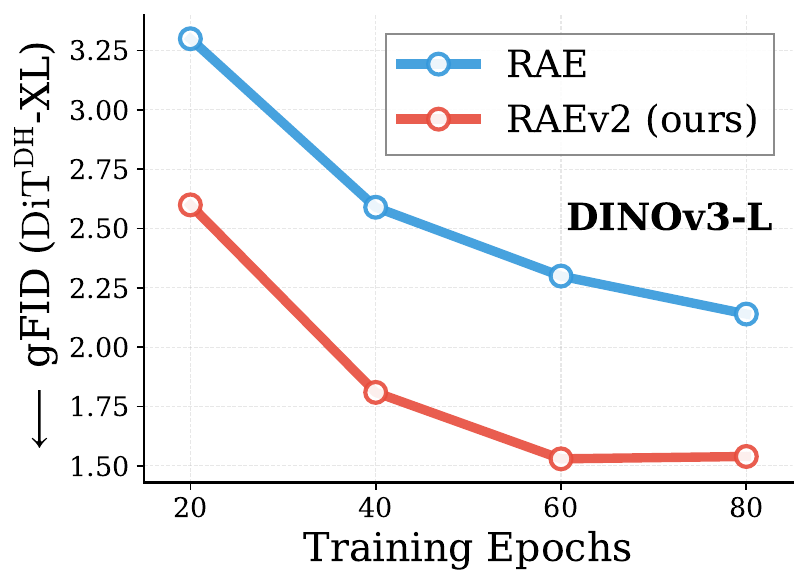}
\end{subfigure}
\begin{subfigure}[b]{0.32\textwidth}
\centering
\includegraphics[width=\linewidth]{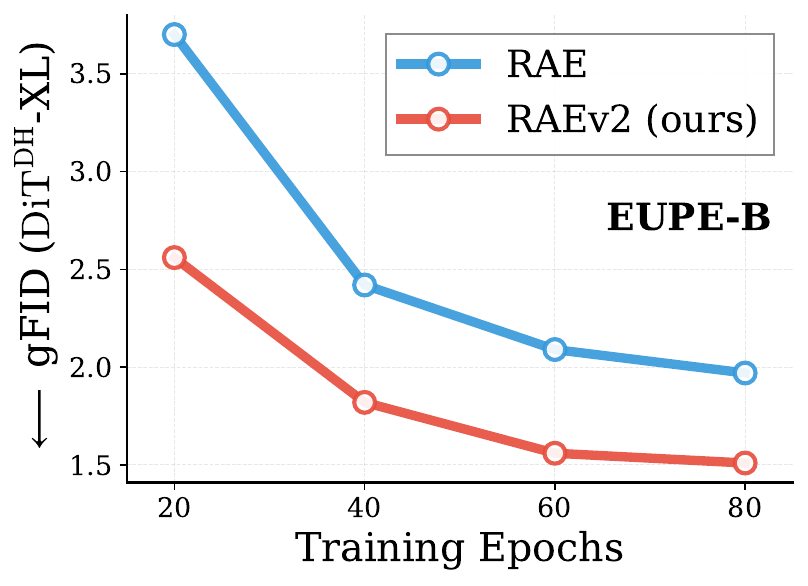}
\end{subfigure}
\hfill
\begin{subfigure}[b]{0.32\textwidth}
\centering
\includegraphics[width=\linewidth]{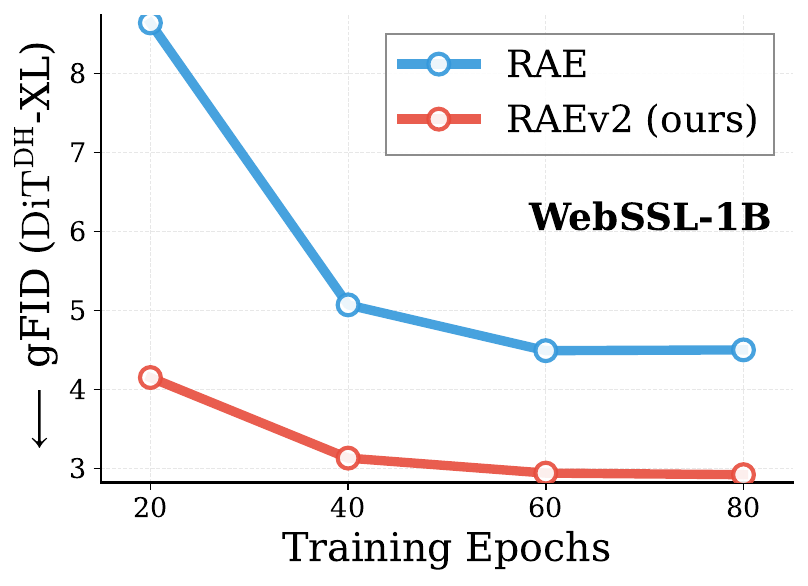}
\end{subfigure}
\hfill
\begin{subfigure}[b]{0.32\textwidth}
\centering
\includegraphics[width=\linewidth]{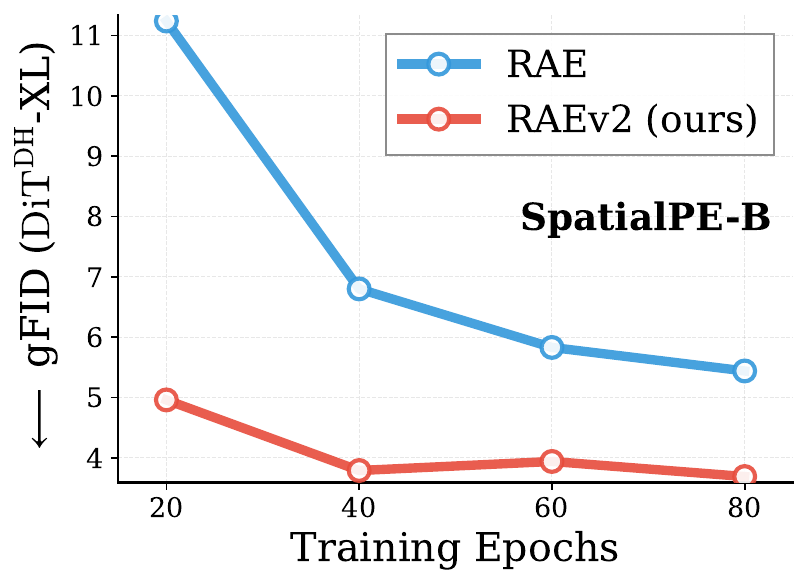}
\end{subfigure}
\vskip -0.05in
\caption{\sethlcolor{black!10}\hl{\textbf{Convergence comparison with original RAE.}} Across DINOv2-B~\cite{dinov2}, DINOv3-B/L~\cite{simeoni2025dinov3}, EUPE-B~\cite{eupe}, WebSSL-1B~\cite{fan2025scaling}, and SpatialPE-B~\cite{bolya2025PerceptionEncoder}, the improved training recipe (RAEv2) consistently leads to faster convergence. All results: DiT$^{DH}$-XL, $K=1$ for RAEv2, 1024 batch-size.}
\label{fig:convergence}
\end{figure*}

\noindent
\sethlcolor{black!10}\hl{\textbf{Convergence speed.}}
Results are shown in \fref{fig:convergence}. We observe that across various vision encoders, RAEv2 consistent improves convergence speed over original RAE.

\begin{wraptable}{r}{0.42\textwidth}
\centering
\vskip -0.2in
\small
\setlength{\tabcolsep}{1.5mm}
\renewcommand{\arraystretch}{1.05}
\begin{tabular}{l c cc}
\toprule
Scale & \#Params & gFID (RAE) $\downarrow$ & gFID (RAEv2) $\downarrow$ \\
\midrule
B  & 165M & 5.48 & \textbf{3.37} \\
L  & 470M & 3.80 & \textbf{2.76} \\
\rowcolor{black!5} XL & 839M & 3.75 & \textbf{2.61} \\
\bottomrule
\end{tabular}
\vskip -0.05in
\caption{{Variation in model scale.} 
}
\label{tab:model_scale}
\vskip -0.4in
\end{wraptable}

\noindent
\sethlcolor{blue!10}\hl{\textbf{Variation in Model scale.}}
We further validate that the gains from RAEv2 transfer across model scales. \tref{tab:model_scale} compares RAE and RAEv2 on DiT$^{DH}$-B, -L, and -XL backbones at 20 epochs: RAEv2 consistently improves generation performance across different scales.

\noindent
\sethlcolor{Apricot!30}\hl{\textbf{Training efficiency.}}
With improved convergence speed of RAEv2 (1.06 gFID in just 80 epochs), we believe that incremental improvements in the gFID metric might provide little signal for practical applications. Instead the training efficiency of a given method, provides much more useful signal for practical applications (e.g., T2I, world models etc \sref{subsec:generalization}). Motivated by the recent speedrun in the language domain \cite{modded_nanogpt_2024}, we therefore report results on $\epfidk$ (number of epochs needed to reach unguided gFID $\le k$). We report results for $k{=}2$ in \tref{tab:fd_eval}. Compared to absolute gFID which shows little variance among various methods, we observe that $\epfidk$ provides a much better signal for measuring training efficiency of a method. Notably, RAE marks a huge jump over prior works reducing $\epfid$ from 480 to 177. RAEv2 further boosts the training efficiency achieving $\epfid$ of just 35 epochs.


\noindent
\sethlcolor{Goldenrod!30}\hl{\textbf{Evaluation with alternate metrics.}}
We also evaluate generation quality with alternate evaluation metrics using recently proposed Representation Fr\'echet Distance (FD$_r$)~\cite{fdr}, which scores sample fidelity in six different feature spaces: Inception, ConvNeXt, DINOv2, MAE, SigLIP, and CLIP. As shown in \tref{tab:fd_eval}, despite training for only 80 epochs, RAEv2 achieves state-of-the-art performance on both FID and FD$_r^6$, surpassing prior baselines that are trained for 800 epochs without any post-training.

\subsection{Impact on Reconstruction Performance}
\label{subsec:rec_gen_analysis}

\noindent
\sethlcolor{purple!15}\hl{\textbf{Qualitative Results.}}
\fref{fig:recon_qualitative} provides qualitative comparisons comparing RAEv2 reconstruction performance with RAE and proprietary VAEs (Flux-VAE, SDVAE, SDXL-VAE). We observe that despite being only trained on ImageNet, RAEv2 shows competitive performance with proprietary VAEs for reconstruction.
We further compare reconstruction quality after using additional data from \cite{raet2i} for training the decoder (encoder remains frozen). We find that RAEv2 shows better reconstruction then SDVAE and SDXL-VAE while performing competitively with Flux-VAE for reconstruction.

\noindent
\begin{wrapfigure}[8]{r}{0.45\textwidth}
\centering
\vskip -0.15in
\includegraphics[width=\linewidth]{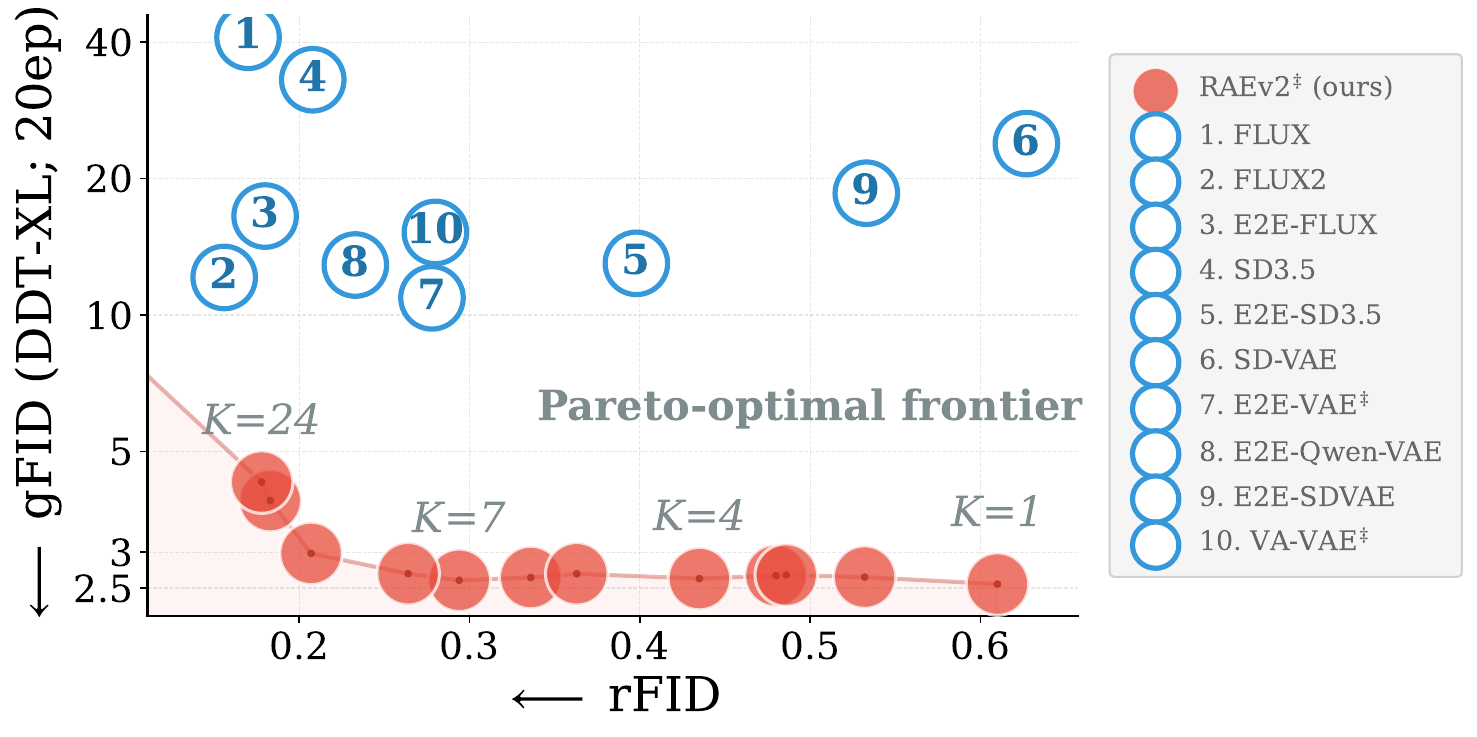}
\vskip -0.05in
\caption{\sethlcolor{BurntOrange!15}\hl{\textbf{Reconstruction-generation trade-off.}}}
\label{fig:rfid_gfid}
\vskip -0.3in
\end{wrapfigure}
\sethlcolor{BurntOrange!15}\hl{\textbf{Reconstruction and generation tradeoff.}}
Results are shown in \fref{fig:rfid_gfid}. RAEv2 achieves Pareto-optimal trade-off between generation (gFID) and reconstruction (rFID)
without encoder finetuning or specialized data (e.g, text) \cite{raet2i}. All results are reported with DINOv3-L encoder, DDT-XL and 20 epochs. Please also see \tref{tab:rec_comparison} for further comparisons.


\begin{table*}[t]
\centering
\vskip -0.4in
\small
\setlength{\tabcolsep}{4pt}
\begin{tabular}{l c cc cccccc c}
\toprule
\multirow{2}{*}{Method} & \multirow{2}{*}{Epochs}
  & \multicolumn{2}{c}{Training Efficiency}
  & \multicolumn{6}{c}{Representation Fr\'echet Distance (FD$_r$)~\cite{fdr} $\downarrow$}
  & \multirow{2}{*}{FD$_r^6$$\downarrow$} \\
\cmidrule(lr){3-4} \cmidrule(lr){5-10}
 & & $\epfid\downarrow$ & gFID$\downarrow$ & Incep. & ConvNeXt & DINOv2 & MAE & SigLIP & CLIP & \\
\midrule
SiT-XL/2~\cite{sit}                   & 800 & $>$800        & 2.12          & 1.26          & 2.02          & 7.89          & 5.62          & 16.14         & 17.69         & 8.44 \\
DDT-XL~\cite{ddt}                     & 800 & --            & 1.26          & 0.75          & 1.02          & 4.26          & 4.11          & 10.16         & 13.86         & 5.70 \\
SiT-XL/2-REPA~\cite{repa}             & 800 & $>$800        & 1.42          & 0.85          & 1.22          & 4.27          & 3.85          &  9.87         & 12.65         & 5.45 \\
LightningDiT~\cite{lgt}               & 800 & $>$800        & 1.42          & 0.85          & 1.09          & 3.76          & 3.02          &  8.47         & 10.21         & 4.57 \\
REG~\cite{reg}                        & 800 & 560           & 1.54          & 0.92          & 1.14          & 3.45          & 3.02          &  8.42         & 10.86         & 4.64 \\
REPA-E~\cite{repae}                   & 800 & 480           & 1.12          & 0.70          & 1.28          & 2.44          & \textbf{2.52} &  5.04         &  6.28         & 3.04 \\
RAE-XL~\cite{rae}                     & 800 & 177           & 1.13          & 0.69          & 1.79          & 2.11          & 3.30          &  3.79         &  7.87         & 3.26 \\
\rowcolor{black!5} \textbf{RAEv2 ($K{=}7$, ours)} & \textbf{80}  & \textbf{35} & \textbf{1.06} & \textbf{0.64} & \textbf{0.77} & \textbf{1.15} & 2.67 & \textbf{2.54} & \textbf{5.21} & \textbf{2.17} \\
\bottomrule
\end{tabular}
\vskip -0.05in
\caption{\sethlcolor{Goldenrod!30}\hl{\textbf{Training efficiency and evaluation under alternative metrics.}}
\textbf{Left:} Training efficiency comparisons reporting $\epfid$ (epochs to reach unguided gFID $\le 2$) and the final guided gFID. Unlike incremental improvements in gFID, $\epfid$ provides a much better signal for training convergence across methods.
\textbf{Right:} Representation Fr\'echet Distance (FD$_r$)~\cite{fdr} computed in six feature spaces (Inception, ConvNeXt, DINOv2, MAE, SigLIP, CLIP), and FD$_r^6$.
Compared to prior works trained for 800 epochs, RAEv2 attains the best $\epfid$, gFID, and FD$_r^6$ in just 80 epochs, without any post-training.
}
\label{tab:fd_eval}
\end{table*}

\section{Generalization to Other Tasks}
\label{sec:generalization}
\label{subsec:generalization}

We further validate the generalization of our improved baseline (RAEv2) on text-to-image generation and navigation world model~\cite{bar2024nwm} tasks. Please refer \sref{sec:appendix_t2i}, \ref{sec:appendix_nwm} for detailed task setup and additional results.

\subsection{Text-to-Image Generation}
\label{subsec:t2i_main}

\begin{figure}[t]
\centering
\includegraphics[width=\linewidth, trim={0 303pt 0 0}, clip]{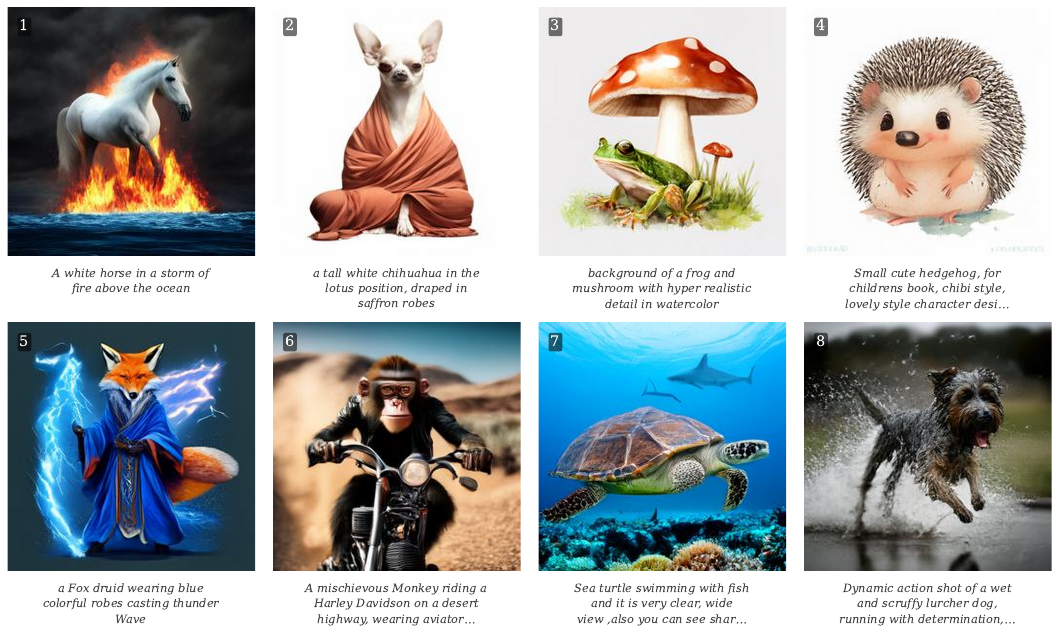}
\caption{\sethlcolor{Plum!10}\hl{\textbf{Text-to-image qualitative samples at 256$\times$256.}} Qualitative samples from RAEv2 (0.9B) trained for 100K iterations with batch size 1024 (equivalent to $\sim$80 epochs on ImageNet at the same batch size), evaluated on MJHQ test set prompts. Additional samples and full prompts are provided in \fref{fig:qualitative_t2i}--\fref{fig:qualitative_t2i_prompts3}.}
\label{fig:qualitative_t2i_main}
\end{figure}

\Paragraph{Training setup.} We first adapt the DiT$^{DH}$-XL backbone for T2I generation. We follow the same incontext architecture from ImageNet experiments (\sref{sec:experiments}), replacing the 8 incontext class-conditional embedding tokens with 256 text-embedding tokens for input captions encoded by Qwen3-0.6B~\cite{qwen2}.
We pretrain on JourneyDB~\cite{journeydb} together with the long-caption and short-caption subsets of BLIP3o~\cite{blip3o} for 150K iterations at batch size 1024, and then finetune on BLIP3o-60k for 50 epochs at the same batch size.

\Paragraph{Evaluation.}
Following~\cite{raet2i}, we report results on GenEval~\cite{geneval}, DPG-Bench~\cite{dpgbench}, and GenAI-Bench~\cite{li2024genai}. All samples are generated with the ODE (Euler) sampler at 50 steps using the EMA model.

\begin{wraptable}{r}{0.5\textwidth}
\centering
\vskip -0.15in
\small
\setlength{\tabcolsep}{1.5mm}
\renewcommand{\arraystretch}{1.05}
\begin{tabular}{l cc cc}
\toprule
 & \multicolumn{2}{c}{Pretraining} & \multicolumn{2}{c}{Finetuning} \\
\cmidrule(lr){2-3} \cmidrule(lr){4-5}
Method & GenEval $\uparrow$ & DPG $\uparrow$ & GenEval $\uparrow$ & DPG $\uparrow$ \\
\midrule
Flux-VAE~\cite{flux} & 41.7          & 77.6          & 78.3          & 79.2          \\
RAE~\cite{rae}       & 58.4          & 80.1          & 81.5          & 80.6          \\
\rowcolor{black!5} RAEv2 (ours) & \textbf{62.4} & \textbf{81.7} & \textbf{82.7} & \textbf{82.3} \\
\bottomrule
\end{tabular}
\vskip -0.05in
\caption{Text-to-image generation.}
\label{tab:t2i_main}
\vskip -0.4in
\end{wraptable}

\Paragraph{Results.}
RAEv2 leads to consistent improvements over Flux-VAE and the original RAE (\tref{tab:t2i_main}). On pretraining, GenEval improves from 41.7 (Flux-VAE) to 62.4 (RAEv2). Similarly after finetuning, RAEv2 reaches 82.7 on GenEval compared 78.3 and 81.5 for Flux-VAE and RAE respectively. \fref{fig:qualitative_t2i_main} shows qualitative visualization of generated samples. The results show strong prompt adherence across diverse subjects despite short training schedule (comparable to 80 epochs on ImageNet). Please see \sref{sec:appendix_t2i} for detailed results.

\subsection{Navigation World Models}
\label{subsec:nwm_main}

\Paragraph{Training setup.}
We further validate our approach for action-conditioned future-frame prediction~\cite{bar2024nwm}, which stress-tests the latent space along two axes: 1) substantially longer conditioning context, and 2) autoregressive rollouts that compound error over time. The model conditions on $N{=}4$ past frames at $256\times 256$ resolution; each frame is encoded by the RAE encoder into a $16\times 16$ patch grid, giving $N \times 256 = 1024$ context tokens (compared to $8$ for class-conditional ImageNet and $256$ for T2I). We additionally append $4$ action tokens (encoding the egocentric action $(\Delta x, \Delta y, \Delta\psi)$) and a Fourier-embedded rollout time token. Following \cite{bar2024nwm}, we train on RECON~\cite{sridhar2023nomad} at 4 FPS, reusing the DiT$^{DH}$-XL backbone and flow-matching recipe from our ImageNet experiments.

\Paragraph{Evaluation.}
Following~\cite{bar2024nwm}, we evaluate predicted frames against ground truth at horizons of $\{1, 2, 4, 8, 16\}$ seconds. Given an FPS of $f$, we obtain the prediction at a target horizon $T$ via $T \cdot f$ autoregressive rollout steps: at each step the model predicts the next frame conditioned on the current sliding window of $N$ context frames and the next ground-truth action, and the predicted RGB is re-encoded and fed back as context. We report FVD~\cite{fvd}, FID~\cite{fid} and LPIPS~\cite{lpips} over the RECON validation split.

\begin{wraptable}{r}{0.3\textwidth}
\centering
\vskip -0.1in
\small
\setlength{\tabcolsep}{2.5mm}
\renewcommand{\arraystretch}{1.05}
\begin{tabular}{l c}
Method & FVD~\cite{fvd} $\downarrow$ \\
\midrule
DIAMOND~\cite{alonso2024diffusion} & 762.73 \\
NWM~\cite{bar2024nwm}              & 200.97 \\
RAE~\cite{rae}                     & 312.01 \\
\rowcolor{black!5} RAEv2 (ours)    & \textbf{105.61} \\
\end{tabular}
\vskip -0.05in
\caption{Video prediction quality upto 16s on RECON \cite{sridhar2023nomad}.}
\label{tab:nwm_fvd}
\vskip -0.4in
\end{wraptable}

\begin{figure}[t]
\begin{minipage}[c]{0.66\linewidth}
\centering
\includegraphics[width=\linewidth]{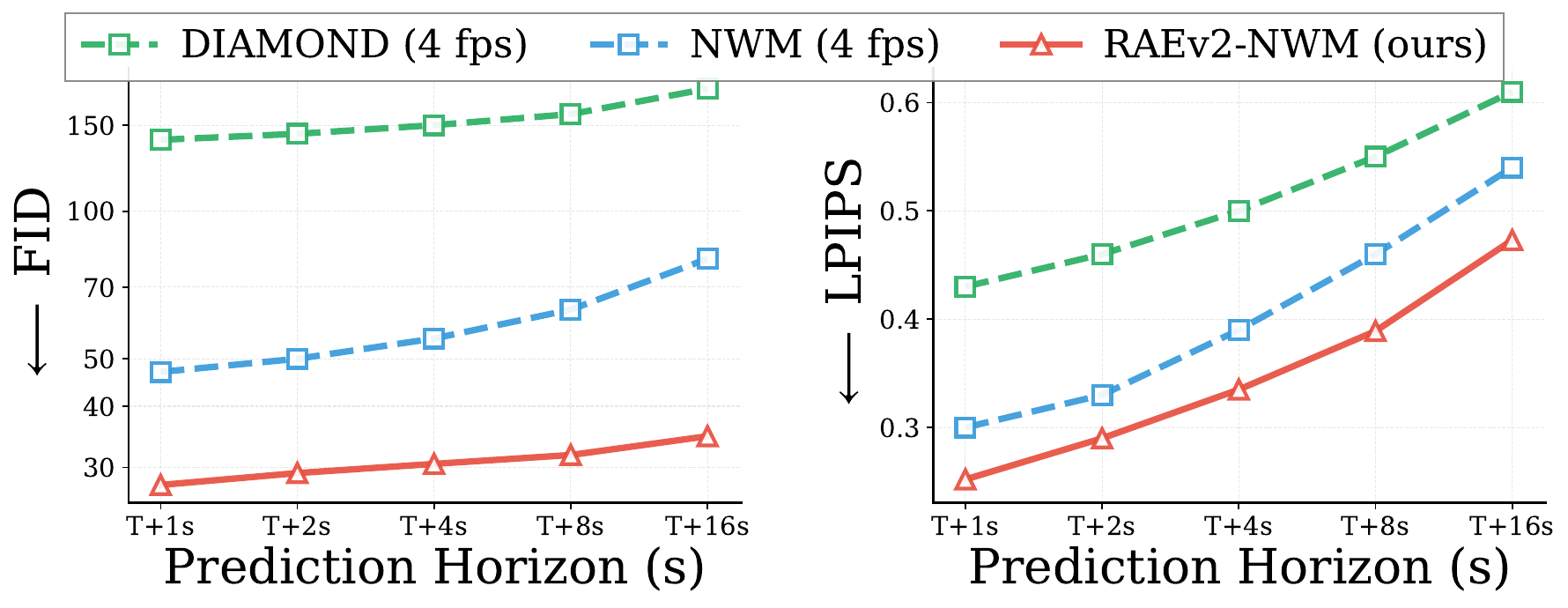}
\end{minipage}\hfill
\begin{minipage}[c]{0.32\linewidth}
\caption{\sethlcolor{black!10}\hl{\textbf{Future state prediction across rollout horizons.}} Comparing generation accuracy and quality of NWM~\cite{bar2024nwm} and DIAMOND~\cite{alonso2024diffusion} at 1 and 4 FPS as function of time, up to 16 seconds of generated video on the RECON dataset.}
\label{fig:nwm_horizon}
\end{minipage}
\end{figure}

\begin{figure}[t]
\centering
\includegraphics[width=\linewidth]{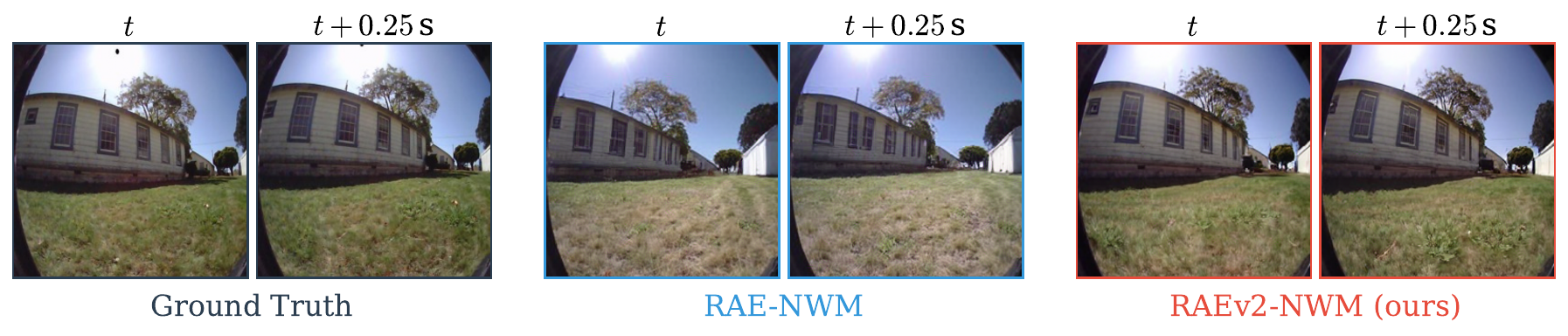}
\caption{\sethlcolor{green!10}\hl{\textbf{Qualitative rollouts with and without the generalized representation autoencoder.}} Consecutive frames at $t$ and $t{+}0.25$s for ground truth, RAE, and RAEv2-NWM (ours, with the generalized RAE of \sref{subsec:generalized_rae}). RAE leads to flickering between consecutive frame predictions (e.g., different number of windows between consecutive frames). In contrast, RAEv2-NWM better retains low-level detail and preserves scene structure across time, which translates into substantially better FVD (\tref{tab:nwm_fvd}).}
\label{fig:nwm_qualitative}
\end{figure}

\Paragraph{Video generation quality.}
On RECON \cite{sridhar2023nomad}, RAEv2-NWM achieves an FVD of 105.61, substantially better than DIAMOND (762.73), NWM (200.97), and RAE (312.01) (\tref{tab:nwm_fvd}). The same ordering holds at every horizon from 1 to 16 seconds on both FID and LPIPS (\fref{fig:nwm_horizon}). Furthermore, we observe that qualitative rollouts also exhibit much less flickering between consecutive frames (\fref{fig:nwm_qualitative}). 

\Paragraph{Importance of generalized representation autoencoders.}
A large fraction of these gains comes from the generalized RAE formulation (\sref{subsec:generalized_rae}), which aggregates the encoder's last $K$ layers rather than relying on the final layer alone. Earlier layers retain low-level texture and geometry critical for temporally consistent navigation rollouts. This leads to better future-state prediction and video quality across rollout horizons; which translates into the substantially lower FVD (\tref{tab:nwm_fvd}).

\begin{wrapfigure}{r}{0.5\textwidth}
\centering
\vskip -0.15in
\includegraphics[width=\linewidth]{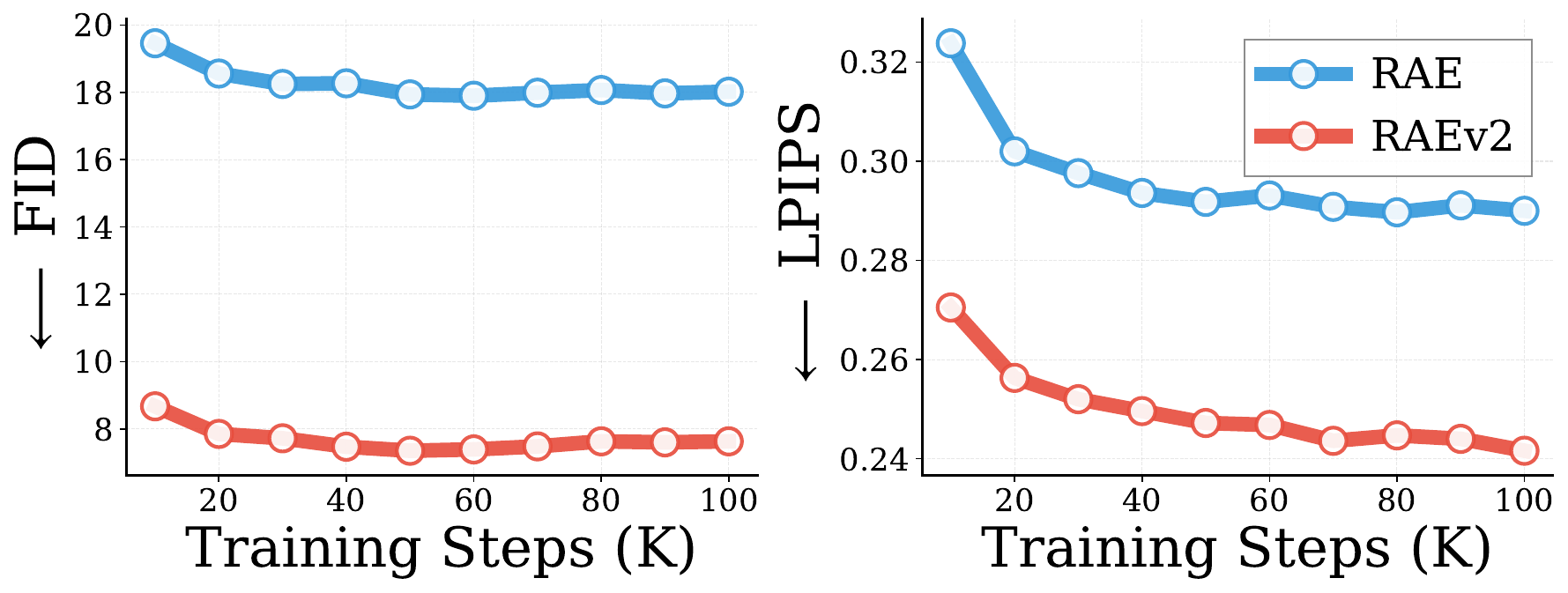}
\vskip -0.05in
\caption{\sethlcolor{blue!10}\hl{\textbf{Convergence speed on RECON validation set.}} FID and LPIPS over training steps for RAE and RAEv2-NWM (ours), evaluated under the single-shot prediction protocol with random target offset $\in [1, 8]$ frames at 4 FPS.}
\label{fig:nwm_convergence}
\vskip -0.2in
\end{wrapfigure}

\Paragraph{Impact on convergence speed.}
\fref{fig:nwm_convergence} shows training curves under the single-shot protocol (no autoregressive rollout) with random target offset $\in [1, 8]$ frames at 4 FPS, i.e.\ predictions $0.25$--$2$ seconds into the future. RAEv2-NWM leads to much faster convergence over RAE; matching RAE's final FID within the first 10K iterations and converges within $\sim$30K to substantially lower FID (7.5 vs.\ 18.0) and LPIPS (0.24 vs.\ 0.29). This mirrors the speedup observed on ImageNet (\sref{subsec:convergence}), indicating that the improved recipe transfers to navigation world models without modification.

\section{Related Work}
\label{sec:related}

We discuss the most relevant work here and provide more detailed discussion in the appendix.

\Paragraph{Pretrained encoders as latent spaces.}
A growing line of work replaces VAE latents with pretrained vision encoders for diffusion \cite{rae, raet2i, svg, chen2025aligning, fae, maetok, reals, flatdino}. We show that the original RAE recipe can be significantly improved with few simple insights leading to more then 10$\times$ faster convergence.

\Paragraph{Representation alignment} distills pretrained representations to intermediate diffusion layers \cite{repa, irepa, repae, reg, ddt}. \cite{chefer2026self} study a complementary direction internalizes representation learning into flow matching itself allowing the model to learn its own representations.
We study the prevalent assumption \cite{rae, riprepa, chang2026dino} that RAE (using pretrained representation as encoder) eliminates the need for representation alignment (REPA). We find that RAE and REPA work through complementary mechanisms.  Their combination is not only useful but allows stronger encoders to perform better and simplifies guidance with RAE. 


\Paragraph{Reconstruction quality of vision encoders}. \cite{raet2i} tries to improve RAE reconstruction using specialized data (text, faces). \cite{lvrae, psvae, uae, dinotok, chang2026dino, aligntok, rpiae, vfmvae} finetune pretrained encoder itself for reconstruction. We find that frozen vision encoders themselves contain low-level details for reconstruction; achieving pareto-optimal reconstruction generation performance without any additional training.



\section{Conclusion}
\label{sec:conclusion}

We study an improved baseline which simplifies and improves RAE.
We find that frozen vision encoders themselves contain low-level details for reconstruction. Simply aggregating the last $K$ layers leads to pareto-optimal reconstruction-generation performance. We next 
perform large-scale empirical analysis showing that RAE and REPA exhibit complementary working mechanisms. 
Their combination is not only useful but also simplifies guidance with RAE. Furthermore it enables stronger representations (e.g., DINOv3-L) which excel in both spatial and global performance to also give better generation performance. Overall, RAEv2 achieves 10$\times$ faster convergence over RAE, improves reconstruction, and achieves state-of-art gFID and FDr$^6$ in just 80 epochs without any post-training. We hope our work provides useful insights for practical adoption of representation autoencoders.


\clearpage
{
    \small
    \bibliographystyle{plainnat}
    \bibliography{main}
}

\appendix
\newpage

\section{Implementation Details}
\label{sec:appendix_implementation}

We provide detailed implementation configurations for reproducibility. \tref{tab:hyperparameters} summarizes all hyperparameters for both class-conditional ImageNet and text-to-image experiments.

\begin{table*}[t]
\centering
\scriptsize
\renewcommand{\arraystretch}{1.15}
\setlength{\tabcolsep}{6pt}
\begin{tabular}{l ccc}
\toprule
Configuration & ImageNet 256$\times$256 & Text-to-Image & World Models \\
\midrule
\rowcolor{black!8} \multicolumn{4}{l}{Architecture} \\
Backbone & DiT$^{DH}$-XL & DiT$^{DH}$-XL & DiT$^{DH}$-XL \\
Encoder blocks / Hidden dim / Heads & 28 / 1440 / 16 & 28 / 1440 / 16 & 28 / 1152 / 16 \\
Decoder blocks / Hidden dim / Heads & 2 / 2048 / 16 & 2 / 2048 / 16 & 2 / 2048 / 16 \\
MLP ratio & 4.0 & 4.0 & 4.0 \\
Patch size (latent) & 1 & 1 & 1 \\
Input channels & 768 & 768 & 768 \\
Conditioning & In-context & In-context & In-context \\
Conditioning tokens & 4 + 8 & 4 + 256 & 1029 \\
Positional embedding & APE + RoPE & APE + RoPE & APE + RoPE \\
Normalization & RMSNorm & RMSNorm & RMSNorm \\
FFN activation & SwiGLU & SwiGLU & SwiGLU \\
\midrule
\rowcolor{black!8} \multicolumn{4}{l}{RAE Encoder} \\
Vision encoder & DINOv3-L & SiGLIP2-B & DINOv3-L \\
Encoder input resolution & 256 & 224 & 256 \\
Encoder patch size & 16 & 16 & 16 \\
Latent shape & $1024 \times 16 \times 16$ & $768 \times 16 \times 16$ & $1024 \times 16 \times 16$ \\
Encoder normalization & Layer norm & Layer norm & Layer norm \\
\midrule
\rowcolor{black!8} \multicolumn{4}{l}{REPA} \\
Target encoder & Same as RAE encoder & Same as RAE encoder & Same as RAE encoder \\
Alignment layer depth & 8 & 8 & 8 \\
Projection type & Linear & Linear & Linear \\
REPA coefficient ($\lambda$) & 1.0 & 1.0 & 1.0 \\
\bottomrule
\end{tabular}
\vskip 0.1in
\caption{\textbf{Architecture and model configurations.} Model architecture, RAE encoder, and REPA settings for class-conditional ImageNet 256$\times$256, text-to-image, and navigation world models. All settings share the same backbone and differ primarily in the conditioning. Continued in \tref{tab:hyperparameters2}.}
\label{tab:hyperparameters}
\end{table*}

\begin{table*}[t]
\centering
\scriptsize
\renewcommand{\arraystretch}{1.15}
\setlength{\tabcolsep}{6pt}
\begin{tabular}{l ccc}
\toprule
Configuration & ImageNet 256$\times$256 & Text-to-Image & World Models \\
\midrule
\rowcolor{black!8} \multicolumn{4}{l}{Training} \\
Dataset & ImageNet-1K & JourneyDB + BLIP3o & RECON \\
Base learning rate & $2 \times 10^{-4}$ & $2 \times 10^{-4}$ & $2 \times 10^{-4}$ \\
Final learning rate & $2 \times 10^{-5}$ & $2 \times 10^{-5}$ & $2 \times 10^{-5}$ \\
LR schedule & Linear decay & Linear decay & Linear decay \\
Warmup epochs / iterations & 25 epochs & 50K iter & 25K iter \\
Warmup decay end (LR reaches final) & 50 epochs & 150K iter & 60K iter \\
Weight decay & 0.0 & 0.0 & 0.0 \\
Global batch size & 1024 & 1024 & 256 \\
Mixed precision & bfloat16 & bfloat16 & bfloat16 \\
Gradient clipping (max norm) & 1.0 & 1.0 & 1.0 \\
EMA decay & 0.9995 & 0.9995 & 0.9995 \\
Training epochs / iterations & 80 & 150K iter (pretrain) + 50 ep (finetune) & 100K iter \\
CFG dropout probability & 0.1 & 0.1 & -- \\
\midrule
\rowcolor{black!8} \multicolumn{4}{l}{Flow Matching} \\
Base prediction type & $x$-prediction & $x$-prediction & $x$-prediction \\
REPA head prediction type & $x$-prediction & $x$-prediction & $x$-prediction \\
Time distribution & Logit-normal & Logit-normal & Logit-normal \\
\midrule
\rowcolor{black!8} \multicolumn{4}{l}{Sampling} \\
Sampler & ODE (Euler) & ODE (Euler) & ODE (Euler) \\
Number of steps & 50 & 50 & 50 \\
Guidance interval & $[0.0, 1.0]$ & $[0.0, 1.0]$ & -- \\
\midrule
\rowcolor{black!8} \multicolumn{4}{l}{Text Conditioning (T2I only)} \\
Text encoder & -- & Qwen3-0.6B & -- \\
Max sequence length & -- & 256 & -- \\
Finetuning dataset & -- & BLIP3o-60k & -- \\
\bottomrule
\end{tabular}
\vskip 0.1in
\caption{\textbf{Training and sampling configurations (continued).} Training hyperparameters, flow matching, sampling, and text conditioning settings. Continuation of \tref{tab:hyperparameters}.}
\label{tab:hyperparameters2}
\end{table*}

\Paragraph{Architecture.}
We use the DDT~\cite{ddt} backbone (DiT$^{DH}$-XL), which consists of a 28-block transformer encoder with hidden dimension 1152 and 16 attention heads, followed by a 2-block DDT decoder with hidden dimension 2048. All layers use RMSNorm, SwiGLU activation in the feed-forward network (MLP ratio 4.0), and rotary positional embeddings (RoPE) combined with absolute positional embeddings (APE). The latent patch size is 1, producing a sequence of $16 \times 16 = 256$ tokens from the encoder output.

\Paragraph{RAE encoder.}
For ImageNet and navigation world model experiments, we use DINOv3-L~\cite{simeoni2025dinov3} as the default encoder. The encoder processes $256 \times 256$ images with patch size 16, producing $16 \times 16 = 256$ patch tokens of dimension 1024, giving a latent representation of shape $1024 \times 16 \times 16$. For text-to-image experiments, we use SiGLIP2-B~\cite{siglip} following~\cite{raet2i}, with the same $16 \times 16$ patch grid and a feature dimension of 768. We discard [CLS] and register tokens and apply layer normalization to the patch outputs. The RAE decoder is pretrained separately for 16 epochs following~\cite{rae} and kept frozen during diffusion training.

\Paragraph{Vision encoders.}
We evaluate pretrained vision encoders across 8 families following~\cite{irepa}: DINOv2~\cite{dinov2}, DINOv3~\cite{simeoni2025dinov3}, WebSSL~\cite{fan2025scaling}, Perception Encoders~\cite{bolya2025PerceptionEncoder}, MoCov3~\cite{mocov3}, CLIP~\cite{clip}, I-JEPA~\cite{ijepa}, and MAE~\cite{mae}. Each encoder is wrapped in a unified interface that extracts patch tokens, discards any [CLS] or register tokens, and applies layer normalization. The full encoder-sweep results, comparing RAE and RAEv2 on every variant, are reported in \tref{tab:encoder_appendix}.

\begin{table}[h!]
\centering
\small
\setlength{\tabcolsep}{3mm}
{
\begin{tabular}{lcccccc}
\toprule
\multirow{2}{*}{Encoder} & \multicolumn{3}{c}{Encoder properties} & \multicolumn{2}{c}{gFID (DiT$^{\text{DH}}$-XL @ 20ep) $\downarrow$} \\
\cmidrule(lr){2-4} \cmidrule(lr){5-6}
 & LP $\uparrow$ & LDS $\uparrow$ & Avg(LP', LDS') $\uparrow$ & RAE & RAEv2 ($k{=}1$) \\
\midrule
MoCov3-B~\cite{mocov3}                & 76.4 & 0.15 & 0.46 & 13.84 & 8.35 \\
CLIP-L~\cite{clip}                    & 84.5 & 0.14 & 0.49 & 7.85  & 4.38 \\
PE-L~\cite{bolya2025PerceptionEncoder}        & 85.5 & 0.14 & 0.50 & 7.06  & 4.09 \\
PE-B~\cite{bolya2025PerceptionEncoder}        & 80.7 & 0.20 & 0.50 & 6.22  & 3.88 \\
WebSSL-1B~\cite{fan2025scaling}       & 84.1 & 0.18 & 0.51 & 8.60  & 4.16 \\
LangPE-L~\cite{bolya2025PerceptionEncoder}    & 83.0 & 0.20 & 0.52 & 5.04  & -- \\
SpatialPE-B~\cite{bolya2025PerceptionEncoder} & 70.8 & 0.33 & 0.52 & 11.24 & 5.04 \\
JEPA-H~\cite{ijepa}                   & 77.5 & 0.33 & 0.55 & 12.46 & 4.48 \\
SpatialPE-L~\cite{bolya2025PerceptionEncoder} & 78.4 & 0.34 & 0.56 & 8.77  & 3.97 \\
DINOv3-B~\cite{simeoni2025dinov3}     & 84.5 & 0.38 & 0.61 & 4.25  & 2.76 \\
DINOv2-B~\cite{dinov2}                & 83.9 & 0.41 & 0.62 & 3.75  & 2.81 \\
\rowcolor{black!5} DINOv3-L~\cite{simeoni2025dinov3} & \textbf{87.0} & \textbf{0.42} & \textbf{0.65} & \textbf{3.30} & \textbf{2.61} \\
\bottomrule
\end{tabular}
}
\vskip 0.1in
\caption{\textbf{Full ablation on choice of pretrained vision encoder.} Extended version of \tref{tab:encoder}. gFID at 20 epochs (DiT$^{\text{DH}}$-XL), sorted by the composite score Avg(LP', LDS'). LP denotes ImageNet linear-probing accuracy (a measure of global semantic quality), and LDS denotes the local-distance similarity score from iREPA~\cite{irepa} (a measure of spatial structure); LP' and LDS' are the min-max normalized values used to form the composite score. RAEv2 consistently improves over RAE across all encoder families. Stronger encoders (e.g., DINOv3-L) which excel at both global and spatial performance achieve the best generation quality. All results are reported without guidance and at batch size 1024.}
\label{tab:encoder_appendix}
\end{table}

\Paragraph{REPA configuration.}
We apply representation alignment at encoder block depth 8 with a single linear projection layer mapping from the transformer hidden dimension (1152) to the target encoder dimension (768). The REPA loss coefficient is set to $\lambda = 0.5$ following~\cite{repa, irepa}. The target encoder is the same as the RAE encoder (self-REPA), as we show in \sref{subsec:rae_repa_orthogonal} that this consistently improves generation across various pretrained encoders.

\Paragraph{Conditioning.}
We replace adaLN-Zero~\cite{dit} with in-context conditioning. The timestep is embedded via Gaussian Fourier features into 4 tokens, and the class label (or text) is embedded into 8 tokens (or up to 256 tokens for T2I). These are concatenated with the image token sequence and processed jointly through self-attention. The DDT decoder strips the conditioning tokens before producing the final output corresponding to the 256 image latent tokens.

\Paragraph{Training.}
We use a base learning rate of $2 \times 10^{-4}$, linearly decayed to $2 \times 10^{-5}$ by epoch 50, with 25 epochs of warmup, and no weight decay. Training uses \texttt{bfloat16} mixed precision with gradient clipping at max norm 1.0. We apply EMA with decay 0.9995 and report all results using the EMA model. All models are trained with global batch size of 1024. 

\Paragraph{Flow matching.}
We use continuous-time flow matching with velocity prediction and logit-normal time sampling following~\cite{sit}. For self-guidance (\sref{subsec:rae_x_prediction}), we convert the model output to $x$-prediction at inference time and apply guidance via the REPA head prediction as defined in Eq.~\ref{eq:self_guidance}.

\Paragraph{Sampling and evaluation.}
We use the ODE solver with Euler discretization for all experiments. We follow the online evaluation protocol from~\cite{jit} and report gFID~\cite{fid} and Inception Score (IS)~\cite{is} on 50K generated images. Following recent work, we additionally report FD$_r$~\cite{fdr} computed across six representation feature spaces (Inception, ConvNeXt, DINOv2, MAE, SigLIP, CLIP), and the geometric mean FD$_r^6$. As a measure of training efficiency, we report $\epfidk$, the number of training epochs to reach unguided gFID $\le k$; we report $k{=}2$. We generate 50 images per class (balanced sampling) following~\cite{rae}.

\Paragraph{Text-to-image.}
We adapt the DiT$^{DH}$-XL backbone for T2I generation. We follow the same incontext architecture from ImageNet experiments (\sref{sec:experiments}), replacing the 8 incontext class-conditional embedding tokens with 256 text-embedding tokens for input captions encoded by Qwen3-0.6B~\cite{qwen2}.
We pretrain on JourneyDB~\cite{journeydb} together with the long-caption and short-caption subsets of BLIP3o~\cite{blip3o} for 150K iterations at batch size 1024, and then finetune on BLIP3o-60k for 50 epochs at the same batch size. We evaluate on GenEval~\cite{geneval}, DPG-Bench~\cite{dpgbench}, and GenAI-Bench~\cite{li2024genai}.

\Paragraph{Navigation world models.}
We use the same DiT$^{DH}$-XL backbone as in the ImageNet and T2I settings, only altering the conditioning tokens to handle navigation inputs. The model conditions on $N=4$ past frames at $256\times 256$ resolution; each frame is encoded by the RAE encoder into a $16\times 16$ patch grid, giving $N \times 256 = 1024$ context tokens. We additionally append $4$ action tokens (encoding the egocentric action $(\Delta x, \Delta y, \Delta\psi)$) and a single Fourier-embedded time token for the rollout offset, for a total of $1029$ conditioning tokens (compared to $8$ for class-conditional ImageNet and $256$ for T2I). Following~\cite{shah2022gnm,shah2023vint,sridhar2023nomad}, we use the RECON~\cite{bar2024nwm,sridhar2023nomad} dataset with the same flow-matching, learning-rate schedule, and EMA recipe as our ImageNet experiments. We train for 100K iterations at batch size 256. For evaluation, following~\cite{bar2024nwm}, we evaluate predicted frames against ground truth at horizons of $\{1, 2, 4, 8, 16\}$ seconds. Given an FPS of $f$, we obtain the prediction at a target horizon of $T$ seconds via $T \cdot f$ autoregressive rollout steps: at each step the model predicts the next frame conditioned on the current sliding window of $N$ context frames and the next ground-truth action, and the predicted RGB is re-encoded and fed back as context. Following \cite{bar2024nwm}, we report FID~\cite{fid}, LPIPS~\cite{lpips} at each horizon, computed over rollout episodes sampled from the held-out RECON~\cite{sridhar2023nomad} validation split. We also report FVD as a measure of video generation quality for autoregressive rollouts upto 16s.


\begin{figure}[t]
\centering
\includegraphics[width=\linewidth]{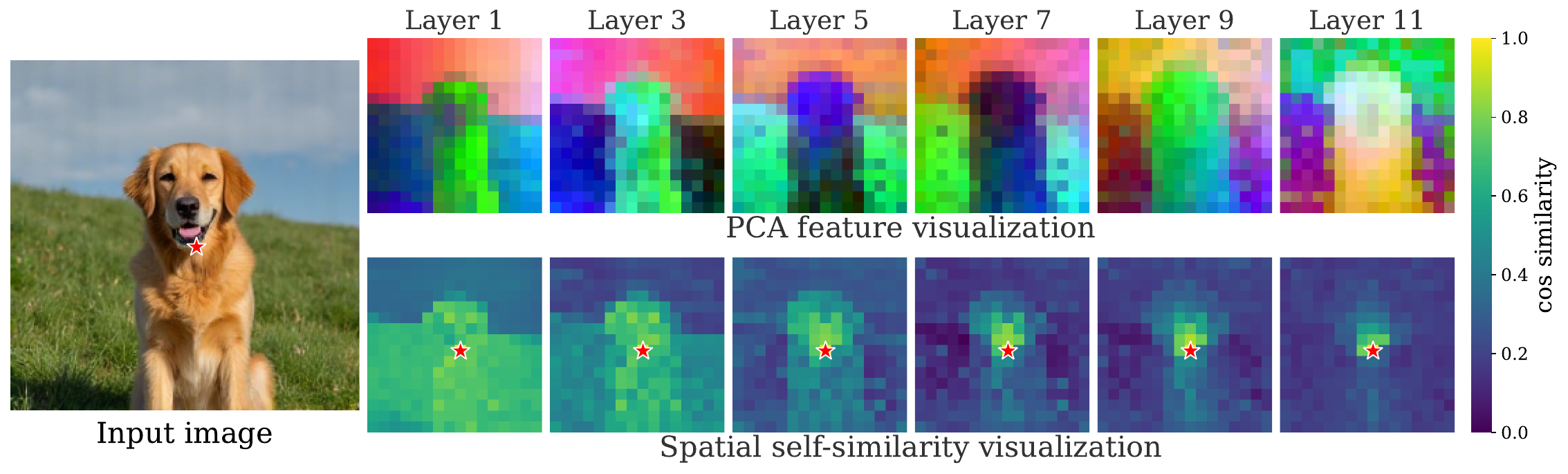}
\caption{\textbf{Different layers of a pretrained encoder provide complementary features.} Aggregating across layers yields richer representations than using the final layer alone.}
\label{fig:per_layer_props}
\end{figure}

\Paragraph{Compute.}
We report results using a $4\times 8$ H100 setup, which trains RAEv2 to gFID 1.06 in roughly $12$ hours, compared to over a week for the original RAE (800 epochs) under the same setup.

\section{Extended Related Work}
\label{sec:appendix_related}

We provide a more comprehensive discussion of related work extending \sref{sec:related}.

\Paragraph{Representation alignment for generation.}
REPA~\cite{repa} aligns intermediate DiT features with pretrained encoders (\eg DINOv2) via a projection head, accelerating convergence. iREPA~\cite{irepa} showed that spatial structure in the alignment target matters more than global information. REPA-E~\cite{repae} extends this to end-to-end VAE tuning. Orthogonally, RAE~\cite{rae} replaces the VAE latent space entirely with pretrained encoder features. LDiT~\cite{ldit} studies the tension between reconstruction and generation in the latent space. A common assumption is that RAE subsumes REPA since both use the same encoder. We find that RAE and REPA exhibit complementary working mechanisms. Their combination is not only useful but also simplifies guidance with RAE.



\Paragraph{Guidance Mechanisms.} Classifier-free guidance (CFG)~\cite{cfg} has become the standard technique for improving sample quality at the cost of diversity, by interpolating between conditional and unconditional predictions. CFG Interval~\cite{cfg_interval} showed that applying guidance only during specific noise levels improves both sample and distribution quality. Autoguidance~\cite{autoguidance} replaced the unconditional model with a weaker conditional model, demonstrating that guidance fundamentally works by contrasting a stronger model against a weaker one rather than requiring unconditional training.

Self-Representation Alignment (SRA)~\cite{sra} showed that diffusion transformers can provide representation guidance by themselves, using intermediate features to steer generation without external models. The dispersive loss~\cite{dispersiveloss} regularizes representations during diffusion training itself to improve generation quality. Internal dynamics guidance~\cite{internalguidance} further explores how a model's own internal representations can substitute for external guidance signals. Recent work on guidance-free generation~\cite{guidancefree} aims to eliminate the need for guidance entirely by incorporating its benefits into training. Our improved self-guidance approach relates to SRA and autoguidance: we show that the REPA prediction head, when combined with $x$-prediction, can serve as an internal guidance signal, avoiding both the unconditional forward pass of CFG and the separate weaker model of autoguidance.

\Paragraph{Representation Learning and Generation.} 
W{\"u}rstchen~\cite{wurstchen} demonstrated that operating in a highly compressed semantic latent space (rather than pixel-level VAE latents) enables efficient large-scale text-to-image synthesis. This insight is closely related to the RAEs of using pretrained encoder features as the diffusion latent space. Several recent works explore how to best construct latent spaces that serve both reconstruction and semantic tasks. FAE~\cite{fae} proposes single-layer adaptation of pretrained features for latent diffusion, showing that minimal fine-tuning of a frozen encoder can yield effective generation latents. MAETok~\cite{maetok} uses a masked autoencoder tokenizer to bridge self-supervised features and discrete token-based generation. FlatDINO~\cite{flatdino} compresses DINOv2 patch features into flatter distributions better suited for diffusion training. ReaLS~\cite{reals} injects semantic priors from pretrained models into the VAE latent space, while SVG~\cite{svg} directly uses frozen DINO features as the generation target.

Unified Latents~\cite{unifiedlatents} jointly trains the encoder, diffusion prior, and decoder with MSE regularization, showing that end-to-end optimization of the full latent pipeline can improve over separately trained components. PS-VAE~\cite{psvae} addresses the tension between semantic richness and pixel-level reconstruction by training representation encoders that excel at both, making them ready for text-to-image generation and editing. These works share a common theme with our approach: the latent space is not merely a compression bottleneck but an active design choice that shapes generation quality, training efficiency, and downstream flexibility. Several works have extended representation alignment beyond static image generation. VideoREPA~\cite{zhang2025videorepa} applies relational alignment with foundation models to video generation, while Geometry Forcing~\cite{wu2025geometry} marries video diffusion with 3D representations. JanusFlow~\cite{ma2025janusflow} unifies multimodal understanding and generation through shared representations and rectified flow.

In this work, we show that pretrained vision encoders themselves have rich representations across different layers. Simply aggregating these features (e.g., through simple addition) enables better generation and reconstruction performance without affecting the understanding performance (measured through linear probing) of the vision encoder.

\section{Additional Results}
\label{sec:appendix_additional_results}

\subsection{Comparisons with original RAE}
\label{sec:appendix_comparisons}

\Paragraph{Additional results on generation-reconstruction performance.}
\tref{tab:rec_comparison} compares RAEv2 against recent representation-based autoencoders (\sref{sec:related}) that target improved reconstruction. All prior works rely on auxiliary losses, encoder fine-tuning, or architectural modifications to the pretrained encoder; in contrast, our generalized RAE formulation (MLS) is strictly training-free, yet simultaneously achieves the best generation quality (at $K{=}7$) and the best reconstruction quality (at $K{=}23$).

\begin{table*}[t]
\centering
\small
\renewcommand{\arraystretch}{1.15}
\begin{tabular}{l c c c c c c c}
\toprule
\multirow{2}{*}{Encoder} & \multirow{2}{*}{\shortstack{Training-free \\ Encoder}} & \multirow{2}{*}{\shortstack{Recon-Gen \\ Tradeoff Control}} & \multirow{2}{*}{Epochs} & \multicolumn{2}{c}{Stage 1} & \multicolumn{2}{c}{Stage 2} \\
\cmidrule(lr){5-6} \cmidrule(lr){7-8}
 & & & & rFID$\downarrow$ & PSNR$\uparrow$ & gFID$\downarrow$ & IS$\uparrow$ \\
\midrule
DINO-Tok~\cite{dinotok}        & \xmark & \xmark & 80 & 0.32  & \textbf{28.54} & 5.94          & 152.6          \\
DINO-SAE~\cite{chang2026dino}  & \xmark & \xmark & 80 & 0.37  & 26.20          & 3.07          & 209.7          \\
VFM-VAE~\cite{vfmvae}          & \xmark & \xmark & 80 & 0.52  & --             & 3.41          & 160.4          \\
AlignTok~\cite{aligntok}       & \xmark & \xmark & 80 & 0.26  & 25.83          & 3.71          & 148.9          \\
RPiAE~\cite{rpiae}             & \xmark & \xmark & 80 & 0.50  & 21.30          & 2.25          & 208.7          \\

RAE~\cite{rae}                 & \cmark & \xmark & 80 & 0.602 & 18.93          & 2.23          & 214.8          \\
\midrule
\rowcolor{black!5}
\textbf{RAEv2 ($K{=}7$, ours)}  & \cmark & \cmark & 80 & 0.29          & 22.57 & \textbf{1.65} & \textbf{228.0} \\
\rowcolor{black!5}
\textbf{RAEv2 ($K{=}23$, ours)} & \cmark & \cmark & 80 & \textbf{0.18} & 27.03 & 3.02          & 206.0          \\
\bottomrule
\end{tabular}
\vskip 0.05in
\caption{\textbf{Reconstruction vs Generation Comparison.} RAEv2 in its generalized form improves both reconstruction and generation performance over recent representation-based autoencoders~\cite{lvrae,psvae,uae,dinotok,chang2026dino,vfmvae,aligntok,rpiae} \emph{without} fine-tuning the pretrained encoder. Furthermore, by simply varying the value of $K$ (the number of last-layer features aggregated), the generalized formulation provides an easy way to control the reconstruction--generation trade-off; on this benchmark RAEv2 achieves the best generation quality at $K{=}7$ and the best reconstruction quality at $K{=}23$.}
\label{tab:rec_comparison}
\end{table*}

\Paragraph{Impact of additional decoder training data for reconstruction.} \tref{tab:decoder_data} reports reconstruction performance for the RAEv2 decoder trained on ImageNet and with additional training data from \cite{raet2i}. Training for longer with more data consistently improves reconstruction. Note: All results are reported with training for only 16 epochs. Training with more data (similar to proprietary
VAEs) and for longer can help further improve reconstruction performance.

\begin{table}[h!]
\centering
\small
\setlength{\tabcolsep}{3mm}
\renewcommand{\arraystretch}{1.1}
\begin{tabular}{l c c c c}
\toprule
Decoder & PSNR $\uparrow$ & SSIM $\uparrow$ & LPIPS $\downarrow$ & rFID $\downarrow$ \\
\midrule
DINOv3-L ($K{=}7$)               & 22.58 & 0.6257 & 0.1531 & 0.299 \\
\;\; + more data                 & 24.18 & 0.6946 & 0.1209 & 0.276 \\
\midrule
DINOv3-L ($K{=}23$)              & 27.04 & 0.8062 & 0.0874 & 0.185 \\
\;\; + more data                 & 29.13 & 0.8625 & 0.0654 & 0.158 \\
\bottomrule
\end{tabular}
\vskip 0.05in
\caption{\textbf{Impact of additional data on RAEv2 decoder training.} Results with and without training on additional data \cite{raet2i} for decoder training. Training for longer with more data consistently improves reconstruction. Note: All results are reported with training for only 16 epochs with frozen pretrained vision encoder. Training with more data (similar to proprietary VAEs) and for longer can help further improve reconstruction performance.}
\label{tab:decoder_data}
\end{table}

\Paragraph{Generalized RAE formulation.} \tref{tab:genrae_formulation_appendix} extends \tref{tab:genrae_formulation} (main paper) with all swept $K \in \{2, 4, 6, 8\}$ values and all five reconstruction/generation metrics (PSNR, SSIM, rFID, gFID, IS). MLS consistently dominates MLR on Stage-2 generation (gFID) across every $K$, while the two methods are essentially tied on the Stage-1 reconstruction metrics.

\begin{table}[h!]
\centering
\setlength{\tabcolsep}{4mm}
\renewcommand{\arraystretch}{1.1}
\newcommand{\gb}{\cellcolor{gray!15}}
\begin{tabular}{c l ccc cc}
\toprule
\multirow{2}{*}{\makecell{Encoder Layers\\(last $K$)}} & \multirow{2}{*}{Method} & \multicolumn{3}{c}{Stage-1 metrics} & \multicolumn{2}{c}{Stage-2 metrics} \\
\cmidrule(lr){3-5} \cmidrule(lr){6-7}
 & & PSNR $\uparrow$ & SSIM $\uparrow$ & rFID $\downarrow$ & gFID $\downarrow$ & IS $\uparrow$ \\
\midrule
\multirow{2}{*}{2}
 & MLR     & 19.72     & 0.509     & 0.570     & 3.085              & --                 \\
 & \gb{}MLS & \gb{}19.44 & \gb{}0.502 & \gb{}0.532 & \gb{}\textbf{2.586} & \gb{}\textbf{243.6} \\
\midrule
\multirow{2}{*}{4}
 & MLR     & 20.86     & 0.558     & 0.425     & 2.954              & --                 \\
 & \gb{}MLS & \gb{}20.50 & \gb{}0.545 & \gb{}0.435 & \gb{}\textbf{2.622} & \gb{}\textbf{230.9} \\
\midrule
\multirow{2}{*}{6}
 & MLR     & 21.97     & 0.607     & 0.342     & 3.118              & --                 \\
 & \gb{}MLS & \gb{}21.92 & \gb{}0.605 & \gb{}0.336 & \gb{}\textbf{2.637} & \gb{}\textbf{223.3} \\
\midrule
\multirow{2}{*}{8}
 & MLR     & 23.36     & 0.669     & 0.268     & 3.580              & --                 \\
 & \gb{}MLS & \gb{}23.30 & \gb{}0.663 & \gb{}0.264 & \gb{}\textbf{2.688} & \gb{}\textbf{220.8} \\
\bottomrule
\end{tabular}
\vskip 0.05in
\caption{\textbf{Full ablation on formulation for generalized RAE.} Extended version of \tref{tab:genrae_formulation}. We compare two parameter-free ways of combining the last $K$ encoder layers (\sref{subsec:generalized_rae}): \textbf{MLS} (multi-layer sum) is a simple addition $\vx = \sum_{\ell} \vz_\ell$; \textbf{MLR} (multi-layer random projection) concatenates the layers and projects back with a fixed random matrix. Encoder is DINOv3-L (24 layers); Stage-1 reports decoder reconstruction (PSNR, SSIM, rFID); Stage-2 reports DiT$^{DH}$-XL training (gFID, IS at 20 epochs). MLS dominates MLR on Stage-2 gFID at every $K$, while the two are essentially tied on Stage-1 reconstruction.}
\label{tab:genrae_formulation_appendix}
\end{table}

\Paragraph{Ablation on guidance mechanism.} \tref{tab:convergence_guidance_appendix} extends \tref{tab:convergence_guidance} (main paper) with the additional Inception Score (IS) column for both $K{=}7$ and $K{=}23$. REPA Guidance achieves the best gFID and IS in both configurations while requiring no separate model and no extra forward pass.

\begin{table}[h!]
\centering
\setlength{\tabcolsep}{4mm}
\renewcommand{\arraystretch}{1.1}
\begin{tabular}{l cc cc}
\toprule
\multirow{2}{*}{Guidance}
  & \multicolumn{2}{c}{RAEv2 ($K{=}7$)}
  & \multicolumn{2}{c}{RAEv2 ($K{=}23$)} \\
\cmidrule(lr){2-3} \cmidrule(lr){4-5}
 & gFID $\downarrow$ & IS $\uparrow$
 & gFID $\downarrow$ & IS $\uparrow$ \\
\midrule
w/o Guidance                          & 1.65          & 228.0          & 3.01          & 206.0          \\
CFG~\cite{cfg}                        & 1.49          & 242.1          & 2.83          & 220.1          \\
Autoguidance (AG)~\cite{autoguidance} & 1.14          & 255.3          & 1.37          & 252.0          \\
\rowcolor{black!5}
REPA Guidance (Ours)                  & \textbf{1.06} & \textbf{255.3} & \textbf{1.25} & \textbf{256.8} \\
\bottomrule
\end{tabular}
\vskip 0.05in
\caption{\textbf{Full ablation on guidance mechanism in RAEv2.} Extended version of \tref{tab:convergence_guidance}, additionally reporting Inception Score (IS). We compare four guidance options for RAEv2 across two encoder-layer aggregation choices ($K{=}7$ and $K{=}23$). REPA Guidance (\sref{subsec:rae_x_prediction}) achieves the best gFID and IS while requiring no additional model (unlike AG) and no extra forward pass (unlike CFG). DiT$^{DH}$-XL backbone with DINOv3-L encoder.}
\label{tab:convergence_guidance_appendix}
\end{table}

\Paragraph{Importance of x-prediction for self-guidance.}
We further ablate the choice of reparameterization, verifying the importance of x-prediction (\sref{subsec:rae_x_prediction}) for self-guidance. \tref{tab:xpred_ablation} compares v-prediction and x-prediction at $K{=}7$ without guidance. We observe that x-prediction, which corresponds to using REPA at intermediate layers, leads to the best generation performance with the generalized RAEv2.

\begin{table}[h!]
\centering
\small
\setlength{\tabcolsep}{4mm}
\renewcommand{\arraystretch}{1.05}
\begin{tabular}{l c c}
\toprule
Parameterization & gFID $\downarrow$ & IS $\uparrow$ \\
\midrule
Internal Guidance \cite{internalguidance}                        & 1.87 & 220.19 \\
\rowcolor{black!5} Internal Guidance w/ x-prediction + REPA-head (ours) & \textbf{1.65} & \textbf{228.00} \\
\bottomrule
\end{tabular}
\vskip 0.05in
\caption{\textbf{Importance of reparameterization to x-prediction with internal-guidance \cite{internalguidance} for RAEv2.} Generation performance at $K{=}7$, 80 epochs and DINOv3-L without guidance. x-prediction (equivalent to using REPA at intermediate layers, \sref{subsec:rae_x_prediction}) outperforms default internal guidance \cite{internalguidance}. Thus, reparameterization to x-prediction is important to achieve the best generation performance with the RAEv2.}
\label{tab:xpred_ablation}
\end{table}

\Paragraph{Impact of generalized RAE on understanding (linear probing).}
A key advantage of RAE is that it provides a unified tokenization for both understanding and generation. While the generalized RAE formulation greatly improves both reconstruction and generation performance (\sref{subsec:generalized_rae}), it is important to understand its impact on the encoder's understanding performance (linear probing). We compare the original DINOv3-L final-layer encoder ($K{=}1$) against the generalized multi-layer-sum (MLS) variants used in RAEv2 ($K{=}7$ and $K{=}23$) in \tref{tab:genrae_lp}. Despite significantly improving reconstruction and generation performance, the generalized RAE formulation does not meaningfully degrade the encoder's understanding performance, as measured by linear probing accuracy on ImageNet.

\begin{table}[h!]
\centering
\small
\setlength{\tabcolsep}{3mm}
\renewcommand{\arraystretch}{1.05}
\begin{tabular}{l c c}
\toprule
Encoder & Feature dim & LP top-1 (\%) $\uparrow$ \\
\midrule
DINOv3-L ($K{=}1$, last layer) & 1024 & 85.39 \\
DINOv3-L MLS ($K{=}2$)         & 1024 & 85.29 \\
DINOv3-L MLS ($K{=}3$)         & 1024 & 85.28 \\
DINOv3-L MLS ($K{=}4$)         & 1024 & 85.15 \\
DINOv3-L MLS ($K{=}5$)         & 1024 & 85.13 \\
DINOv3-L MLS ($K{=}6$)         & 1024 & 85.14 \\
DINOv3-L MLS ($K{=}7$)         & 1024 & 85.10 \\
DINOv3-L MLS ($K{=}8$)         & 1024 & 85.10 \\
DINOv3-L MLS ($K{=}9$)         & 1024 & 85.10 \\
DINOv3-L MLS ($K{=}10$)        & 1024 & 85.12 \\
DINOv3-L MLS ($K{=}23$, full)  & 1024 & 85.24 \\
\bottomrule
\end{tabular}
\vskip 0.05in
\caption{\textbf{Impact of generalized RAE on understanding (linear probing).} Linear probing top-1 accuracy on ImageNet across all $K \in \{1, \dots, 10, 23\}$ for the generalized multi-layer-sum (MLS) variant on DINOv3-L. $K{=}1$ corresponds to the original RAE (final-layer feature). The generalized formulation (\sref{subsec:generalized_rae}) improves reconstruction without meaningfully impacting global semantic performance, enabling unified tokenization for both understanding and generation. All values are computed at 30 epochs of LP training with learning rate $1\times 10^{-2}$; continued training may further improve linear probing scores.}
\label{tab:genrae_lp}
\end{table}

\Paragraph{Evaluation under the Monge Distance.}
Following the recent MIND framework~\cite{mind}, we additionally evaluate RAEv2 against RAE and REPA-E using the Monge Distance, an optimal-transport based alternative to the Fr\'echet Distance. \tref{tab:monge_eval} reports the Representation Monge Distance (MD$_r$) computed across the same six feature spaces used for FD$_r$ in \tref{tab:fd_eval}. RAEv2 attains the best MD$_r$ in five of six feature spaces in just 80 epochs, further corroborating the strong results under alternative evaluation metrics.

\begin{table*}[t]
\centering
\small
\setlength{\tabcolsep}{4pt}
\begin{tabular}{l c cccccc}
\toprule
\multirow{2}{*}{Method} & \multirow{2}{*}{Epochs}
  & \multicolumn{6}{c}{Representation Monge Distance (MD$_r$)~\cite{mind} $\downarrow$} \\
\cmidrule(lr){3-8}
 & & Incep. & ConvNeXt & DINOv2 & MAE & SigLIP & CLIP \\
\midrule
REPA-E~\cite{repae}  & 800 & 1.112 \scriptsize{$\pm$0.08} & 56.63 \scriptsize{$\pm$1.69} & 26.82 \scriptsize{$\pm$0.55} & 0.196 \scriptsize{$\pm$0.01} &  4.44 \scriptsize{$\pm$0.12} & 44.75 \scriptsize{$\pm$0.14} \\
RAE-XL~\cite{rae}    & 800 & \textbf{0.808 \scriptsize{$\pm$0.04}} & 70.29 \scriptsize{$\pm$1.87} & 19.70 \scriptsize{$\pm$0.32} & 0.230 \scriptsize{$\pm$0.01} &  2.96 \scriptsize{$\pm$0.17} & 68.46 \scriptsize{$\pm$1.18} \\
\rowcolor{black!5} \textbf{RAEv2 ($K{=}7$, ours)} & \textbf{80}  & 0.997 \scriptsize{$\pm$0.04} & \textbf{31.71 \scriptsize{$\pm$0.58}} & \textbf{7.27 \scriptsize{$\pm$0.20}} & \textbf{0.133 \scriptsize{$\pm$0.00}} & \textbf{1.71 \scriptsize{$\pm$0.08}} & \textbf{41.68 \scriptsize{$\pm$2.66}} \\
\bottomrule
\end{tabular}
\caption{\textbf{Evaluation under the Monge Distance.} Following~\cite{mind}, we additionally evaluate methods using the Monge Distance~\cite{mind} as an alternative to the Fr\'echet Distance. Analogous to FD$_r$ \cite{fdr}, we report the Representation Monge Distance (MD$_r$) computed in six feature spaces (Inception, ConvNeXt, DINOv2, MAE, SigLIP, CLIP). Compared to prior baselines trained with 800 epochs, RAEv2 attains the best MD$_r$ with different feature spaces in just 80 epochs, without any post-training. All results with 50K evaluation samples.}
\label{tab:monge_eval}
\end{table*}

\subsection{Text-to-Image Generation}
\label{sec:appendix_t2i}
\label{sec:appendix_generalization}

\Paragraph{Training setup.}
We pretrain a text-to-image model from scratch on JourneyDB~\cite{journeydb} together with the long-caption and short-caption subsets of BLIP3o~\cite{blip3o}, for 150K iterations at a global batch size of 1024. Following~\cite{raet2i}, we use SiGLIP2-B~\cite{siglip} as the RAE encoder and adapt the DiT$^{DH}$-XL backbone for text-conditioning. Text captions are encoded by Qwen3-0.6B~\cite{qwen2} with a maximum sequence length of 256 tokens. Optimization mirrors the ImageNet recipe (lr $2\times 10^{-4}$ linearly decayed to $2\times 10^{-5}$, \texttt{bfloat16}, EMA decay 0.9995). We then finetune on the BLIP3o-60k subset for 50 epochs at the same batch size.

\Paragraph{Evaluation.}
Following~\cite{raet2i}, we report results on GenEval~\cite{geneval}, DPG-Bench~\cite{dpgbench}, and GenAI-Bench~\cite{li2024genai}, covering compositional, dense-prompt, and human-preference axes. Samples are generated with the ODE (Euler) sampler at 50 steps using the EMA model.

\noindent
\sethlcolor{Plum!10}\hl{\textbf{Pretraining.}} Results are shown in Tab.~\ref{tab:t2i}. We observe that as compared to widely used Flux-VAE \cite{flux}, the use of representation autoencoders leads to significant improvements for text-to-image generation. Furthermore, using the improved training recipe leads to even further gains across all evaluation metrics. For instance, while Flux-VAE and RAE lead to a GenEval score of 41.7 and 58.4 respectively, the use of improved baseline RAEv2 leads to better performance with GenEval score of 62.4.

\noindent
\sethlcolor{cyan!10}\hl{\textbf{Finetuning.}}
Following \cite{raet2i}, we also perform finetuning of our pretrained model using the 60k finetuning dataset from BLIP3o \cite{blip3o}. We use a batch size of 1024 and 50 epochs for finetuning. Similar to findings of \cite{raet2i}, we find that this helps significantly increase the performance to 82.7 on GenEval with RAEv2. Furthermore, while finetuning reduces the gap between various methods, RAEv2 still shows improved performance over original RAE and Flux-VAE.


\begin{table}[h!]
\centering
{
\begin{tabular}{llcccc}
\toprule
Method & Model & Params & GenEval$\uparrow$ & GenAI-Bench$\uparrow$ & DPG-Bench$\uparrow$ \\
\midrule
\rowcolor{Plum!10} \multicolumn{6}{l}{Pretraining} \\
\arrayrulecolor{black!30}\midrule
Flux-VAE \cite{flux} & DiT$^{\text{DH}}$-XL & 0.9B & 41.7 & 57.3 & 77.6 \\
RAE \cite{rae} & DiT$^{\text{DH}}$-XL & 0.9B & 58.4 & 63.2 & 80.1 \\
\rowcolor{black!5} RAEv2 & DiT$^{\text{DH}}$-XL & 0.9B & \textbf{62.4} & \textbf{63.8} & \textbf{81.7} \\
\arrayrulecolor{black}\midrule
\rowcolor{cyan!10} \multicolumn{6}{l}{Finetuning} \\
\arrayrulecolor{black!30}\midrule
Flux-VAE \cite{flux} & DiT$^{\text{DH}}$-XL & 0.9B & 78.3 & 63.9 & 79.2 \\
RAE \cite{rae} & DiT$^{\text{DH}}$-XL & 0.9B & 81.5 & 67.2 & 80.6 \\
\rowcolor{black!5} RAEv2 & DiT$^{\text{DH}}$-XL & 0.9B & \textbf{82.7} & \textbf{68.0} & \textbf{82.3} \\
\arrayrulecolor{black}\bottomrule
\end{tabular}
}
\vskip 0.1in
\caption{\sethlcolor{Plum!10}\hl{\textbf{Text-to-image generation.}} Results comparing proposed RAEv2 with original RAE \cite{rae} and Flux-VAE \cite{flux}. Results for pretraining are reported at 150K steps with batch-size of 1024 and JourneyDB, long-caption and short-caption subsets from BLIP3o pretraining subset \cite{blip3o}. For finetuning similar to \cite{raet2i}, we use the 60k subset from \cite{blip3o}, and 1024 batchsize. Across all settings, we observe that RAEv2 leads to faster training over original RAE and Flux-VAE.}
\label{tab:t2i}
\end{table}

\subsection{Navigation World Models}
\label{sec:appendix_nwm}

We follow the navigation world modeling setup of NWM~\cite{bar2024nwm}. In this setting, the model is conditioned on the last $N=4$ egocentric video frames together with an action sequence, and is trained to predict the next frame in the trajectory. At inference time, the model rolls out future frames \emph{autoregressively}: at each step, the predicted frame is fed back into the context window so that long-horizon predictions can be produced from a short history.

\Paragraph{Training setup.}
We use the same DiT$^{DH}$-XL backbone as in the previous sections, only altering the conditioning tokens to handle navigation inputs. The model conditions on $N=4$ past frames at $256\times 256$ resolution; each frame is encoded by the RAE encoder into a $16\times 16$ patch grid, giving $N \times 256 = 1024$ context tokens. We additionally append $4$ action tokens (encoding the egocentric action $(\Delta x, \Delta y, \Delta\psi)$) and a single Fourier-embedded time token for the rollout offset, for a total of $1029$ conditioning tokens (compared to $8$ for class-conditional ImageNet and $256$ for T2I). Following~\cite{shah2022gnm,shah2023vint,sridhar2023nomad}, we use the RECON~\cite{bar2024nwm,sridhar2023nomad} dataset with the same flow-matching, learning-rate schedule, and EMA recipe as our ImageNet experiments. We train for 100K iterations at a batch size of 256, on the same $4\times 8$ H100 setup used for the ImageNet experiments.

\Paragraph{Evaluation.}
Following~\cite{bar2024nwm}, we evaluate predicted frames against ground truth at horizons of $\{1, 2, 4, 8, 16\}$ seconds. Given an FPS of $f$, we obtain the prediction at a target horizon of $T$ seconds via $T \cdot f$ autoregressive rollout steps: at each step the model predicts the next frame conditioned on the current sliding window of $N$ context frames and the next ground-truth action, and the predicted RGB is re-encoded and fed back as context. We report FID~\cite{fid}, LPIPS~\cite{lpips}, PSNR, and DreamSim~\cite{fu2023dreamsim} at each horizon, computed over rollout episodes sampled from the RECON validation split.

\noindent
\sethlcolor{black!10}\hl{\textbf{Future state prediction and synthesis.}}
Across rollout horizons RAEv2-NWM produces noticeably more accurate and temporally stable predictions than DIAMOND, NWM, and the RAE baseline. Quantitatively, RAEv2-NWM achieves an FVD of 105.61 on the RECON validation set, compared to 762.73 for DIAMOND, 200.97 for NWM, and 312.01 for RAE (\tref{tab:nwm_fvd_appendix}); the same ordering holds at every horizon from 1 to 16 seconds on both FID and LPIPS (\fref{fig:nwm_horizon}). Qualitatively, the rollouts also exhibit much less flickering between consecutive frames (\fref{fig:nwm_qualitative}).

\begin{table}[h!]
\centering
\setlength{\tabcolsep}{8pt}
\renewcommand{\arraystretch}{1.15}
\begin{tabular}{l c c c c}
\toprule
                                & DIAMOND~\cite{alonso2024diffusion} & NWM~\cite{bar2024nwm} & RAE~\cite{rae} & RAEv2 (ours) \\
\midrule
\#Params               & 1B     & 1B     & 622M   & 622M            \\
FVD~\cite{fvd} $\downarrow$ & 762.73 & 200.97 & 312.01 & \textbf{105.61} \\
\bottomrule
\end{tabular}
\vskip 0.05in
\caption{\textbf{Video prediction quality on RECON~\cite{sridhar2023nomad}.} FVD computed over autoregressive rollouts up to 16s. Reference values for DIAMOND and NWM are from~\cite{bar2024nwm}.}
\label{tab:nwm_fvd_appendix}
\end{table}


\noindent
\sethlcolor{green!10}\hl{\textbf{Importance of generalized representation autoencoders.}}
A large fraction of these gains comes from using the generalized RAE formulation (\sref{subsec:generalized_rae}), which aggregates the encoder's last $K$ layers rather than relying on the final layer alone. The earlier layers retain low-level texture and geometry that are critical for temporally consistent navigation rollouts. As a result, the generalized formulation converges substantially faster during training (\fref{fig:nwm_convergence}; reaching the RAE baseline's final error within roughly 10K iterations), and produces better future-state prediction and video quality across rollout horizons, translating into the substantially lower FVD reported in \tref{tab:nwm_fvd}.

\noindent
\sethlcolor{blue!10}\hl{\textbf{Impact on convergence speed.}}
\fref{fig:nwm_convergence} shows training curves on RECON under the online single-shot protocol with random offset $\in [1, 8]$ frames at 4 FPS, i.e. predictions $0.25$--$2$ seconds into the future. RAEv2-NWM converges within $\sim$30K iterations to noticeably lower FID and LPIPS than the RAE baseline (FID 7.5 vs.\ 18.0, LPIPS 0.24 vs.\ 0.29), and matches the RAE baseline's final FID within the first 10K iterations. This mirrors the speedup we observe on ImageNet (\sref{subsec:convergence}) and indicates that the improved recipe transfers to navigation world models without modification.

\section{Qualitative Results}
\label{sec:appendix_qualitative}

%

\begin{figure*}[t]
\centering
\includegraphics[width=\textwidth]{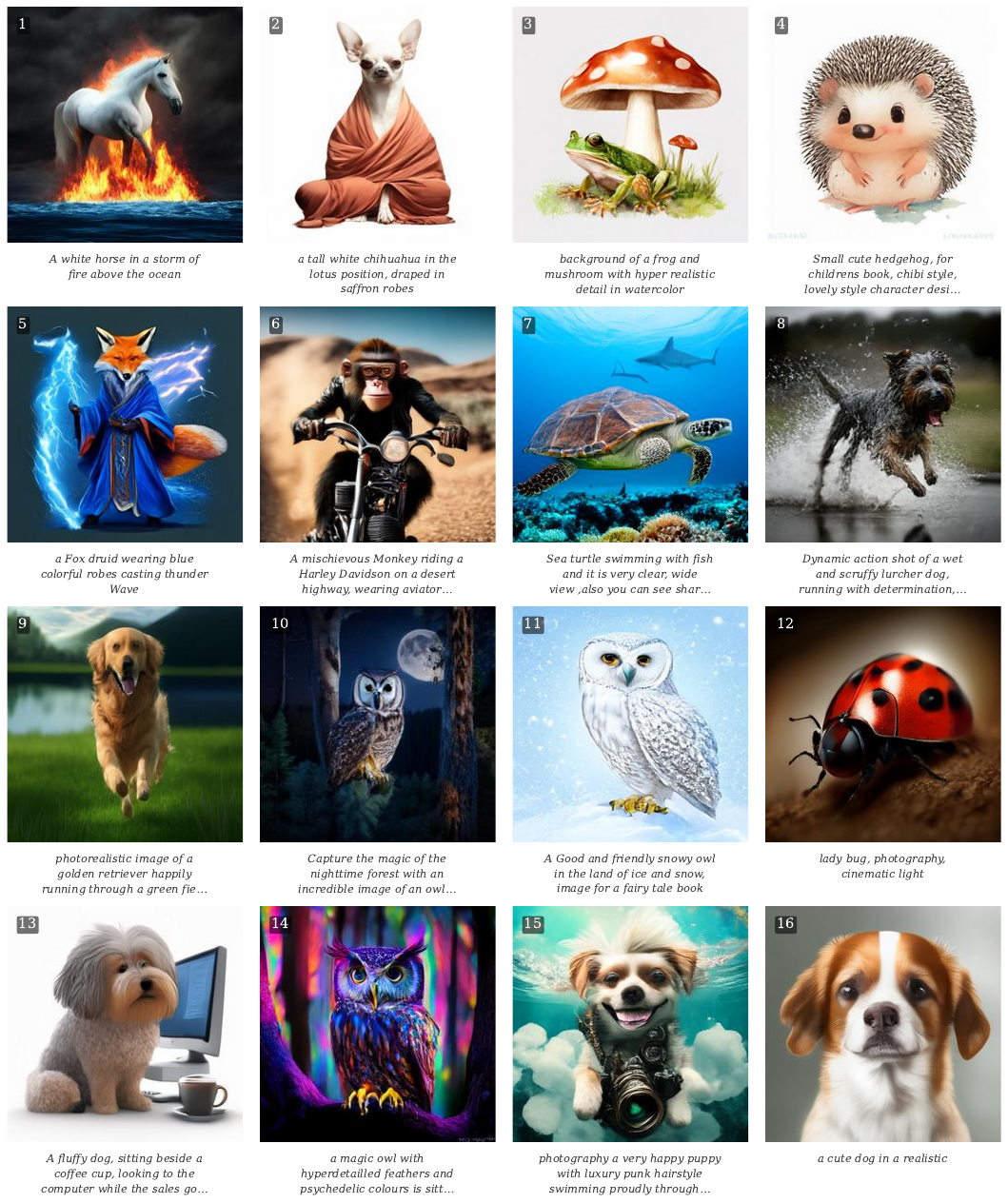}
\vskip -0.05in
\caption{\textbf{Text-to-image qualitative examples at 256$\times$256 resolution (1/3).} RAEv2 (0.9B) trained for 100K iterations with batch size 1024, evaluated on MJHQ test set prompts. Corresponding prompts are listed in \fref{fig:qualitative_t2i_prompts}.}
\label{fig:qualitative_t2i}
\end{figure*}
\begin{figure*}[t]
\centering
\includegraphics[width=\textwidth]{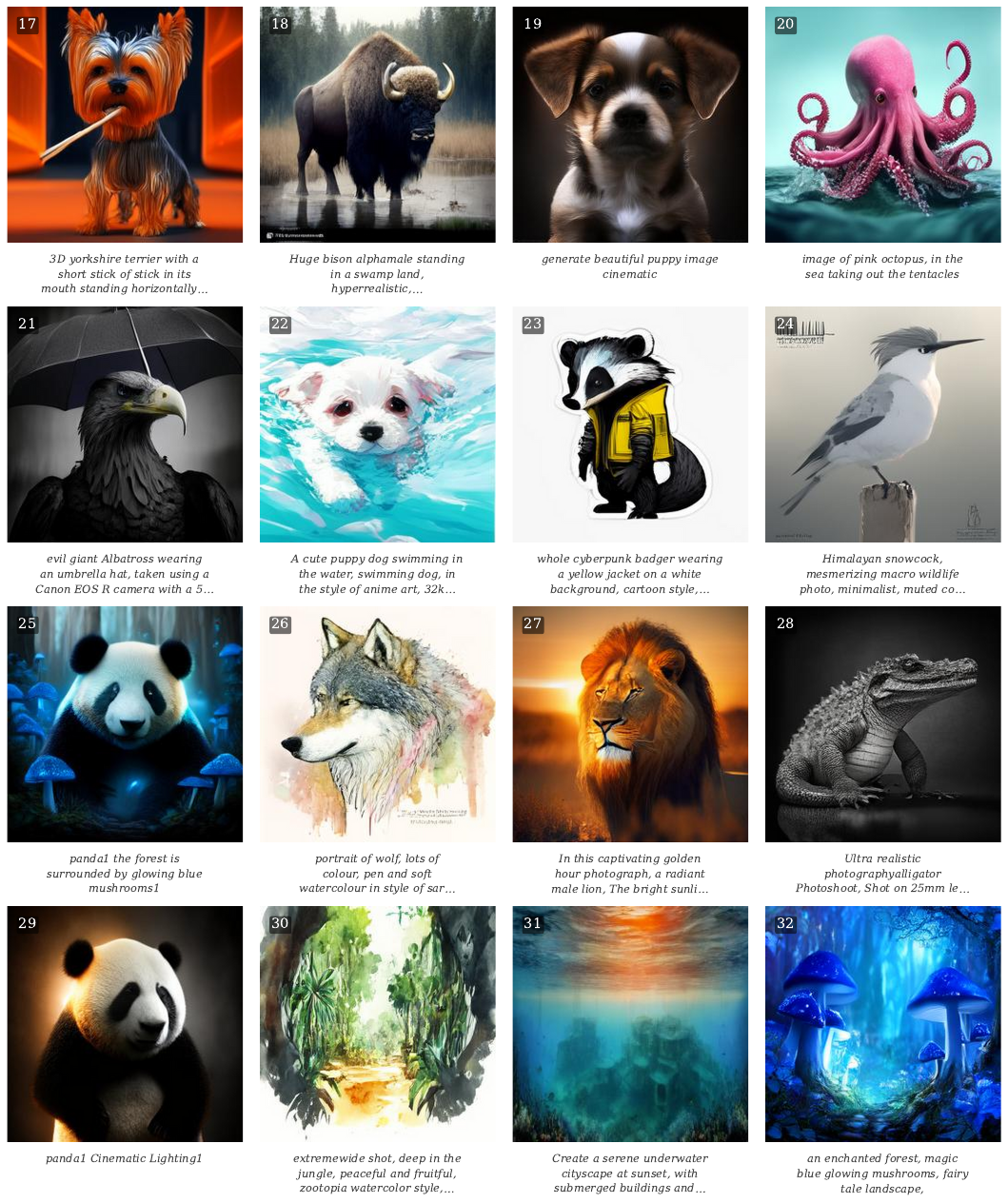}
\vskip -0.05in
\caption{\textbf{Text-to-image qualitative examples at 256$\times$256 resolution (2/3).} RAEv2 (0.9B) trained for 100K iterations with batch size 1024, evaluated on MJHQ test set prompts. Corresponding prompts are listed in \fref{fig:qualitative_t2i_prompts}.}
\label{fig:qualitative_t2i2}
\end{figure*}
\begin{figure*}[t]
\centering
\includegraphics[width=\textwidth]{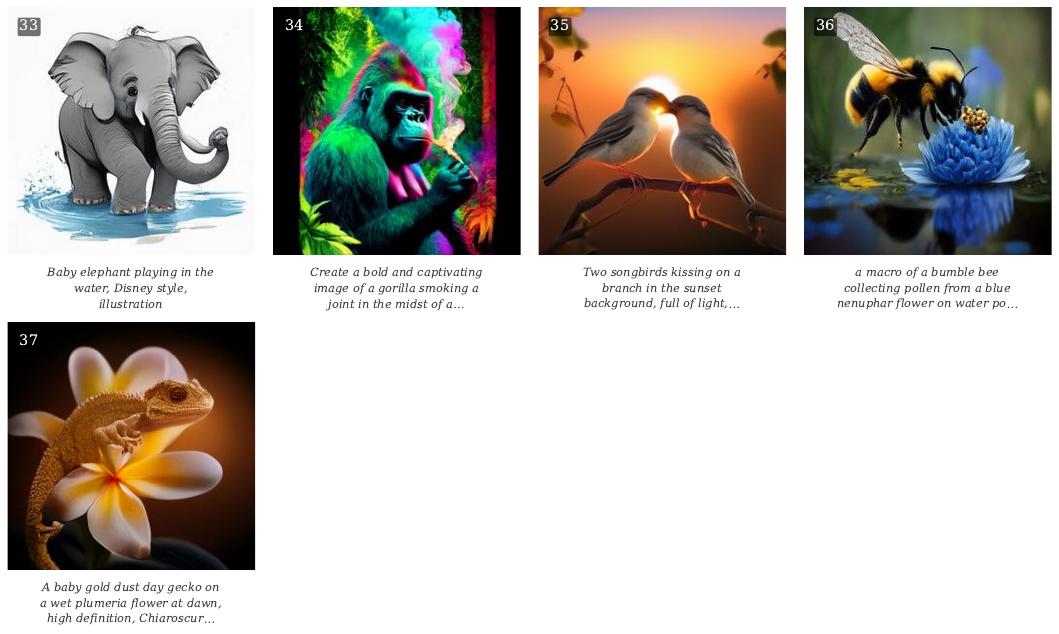}
\vskip -0.05in
\caption{\textbf{Text-to-image qualitative examples at 256$\times$256 resolution (3/3).} RAEv2 (0.9B) trained for 100K iterations with batch size 1024, evaluated on MJHQ test set prompts. Corresponding prompts are listed in \fref{fig:qualitative_t2i_prompts}.}
\label{fig:qualitative_t2i3}
\end{figure*}
\begin{figure*}[t]
\centering
\includegraphics[width=\textwidth]{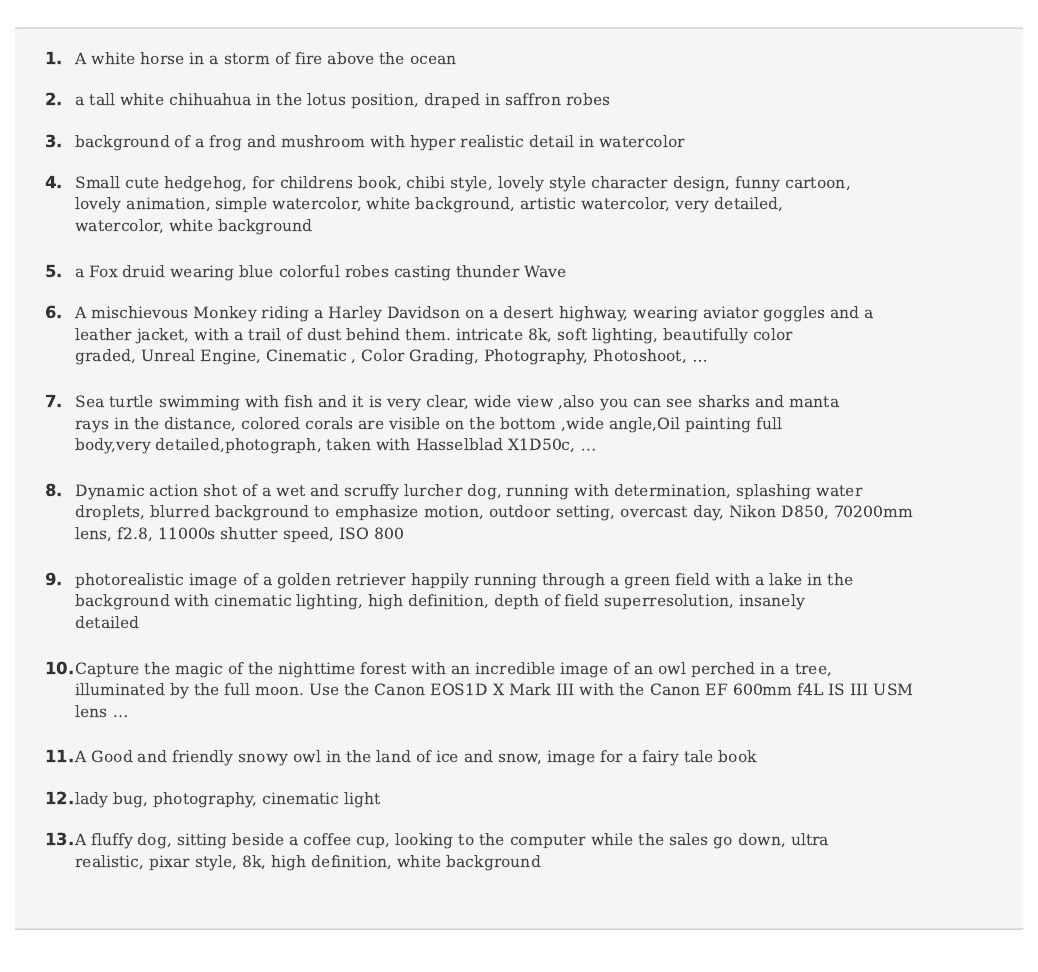}
\vskip -0.05in
\caption{\textbf{Text prompts for T2I qualitative samples (1/3).} Prompts corresponding to the generated images in \fref{fig:qualitative_t2i}--\fref{fig:qualitative_t2i3}.}
\label{fig:qualitative_t2i_prompts}
\end{figure*}
\begin{figure*}[t]
\centering
\includegraphics[width=\textwidth]{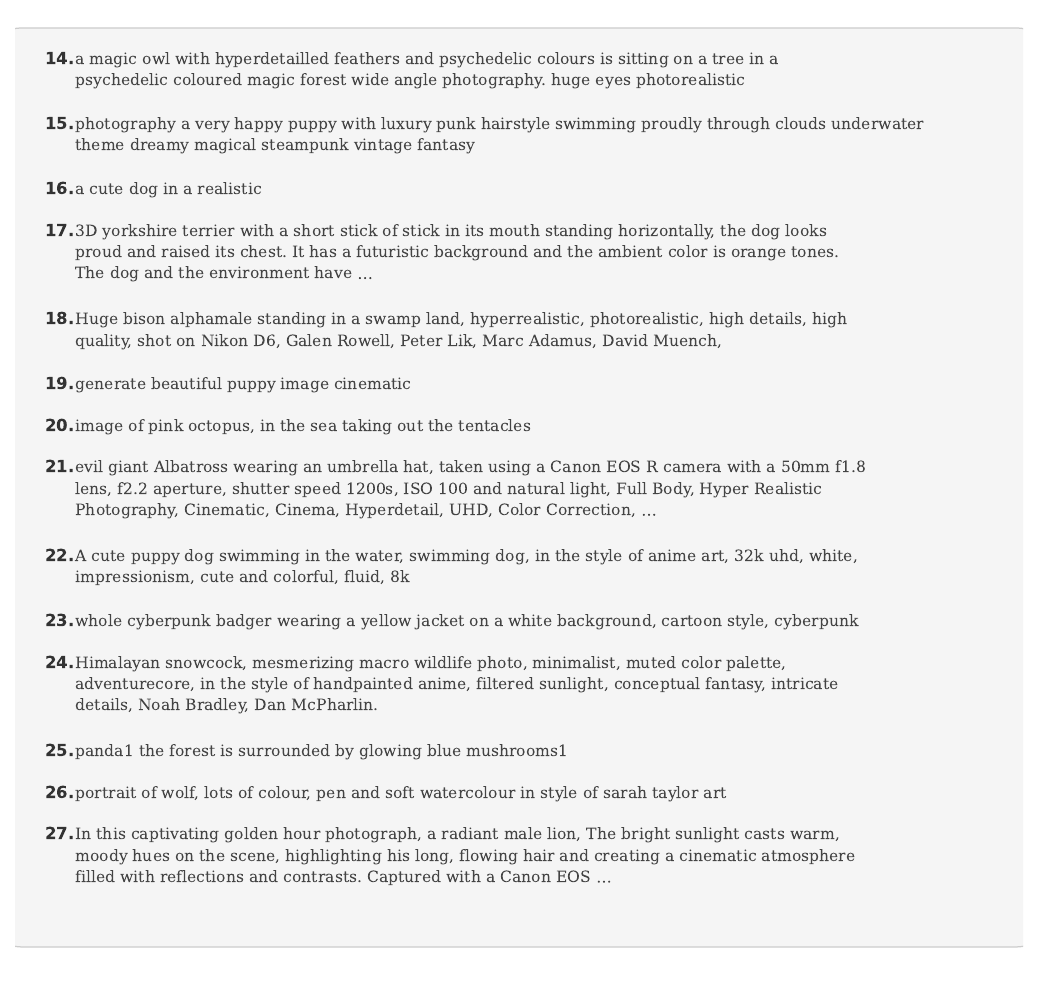}
\vskip -0.05in
\caption{\textbf{Text prompts for T2I qualitative samples (2/3).}}
\label{fig:qualitative_t2i_prompts2}
\end{figure*}
\begin{figure*}[t]
\centering
\includegraphics[width=\textwidth]{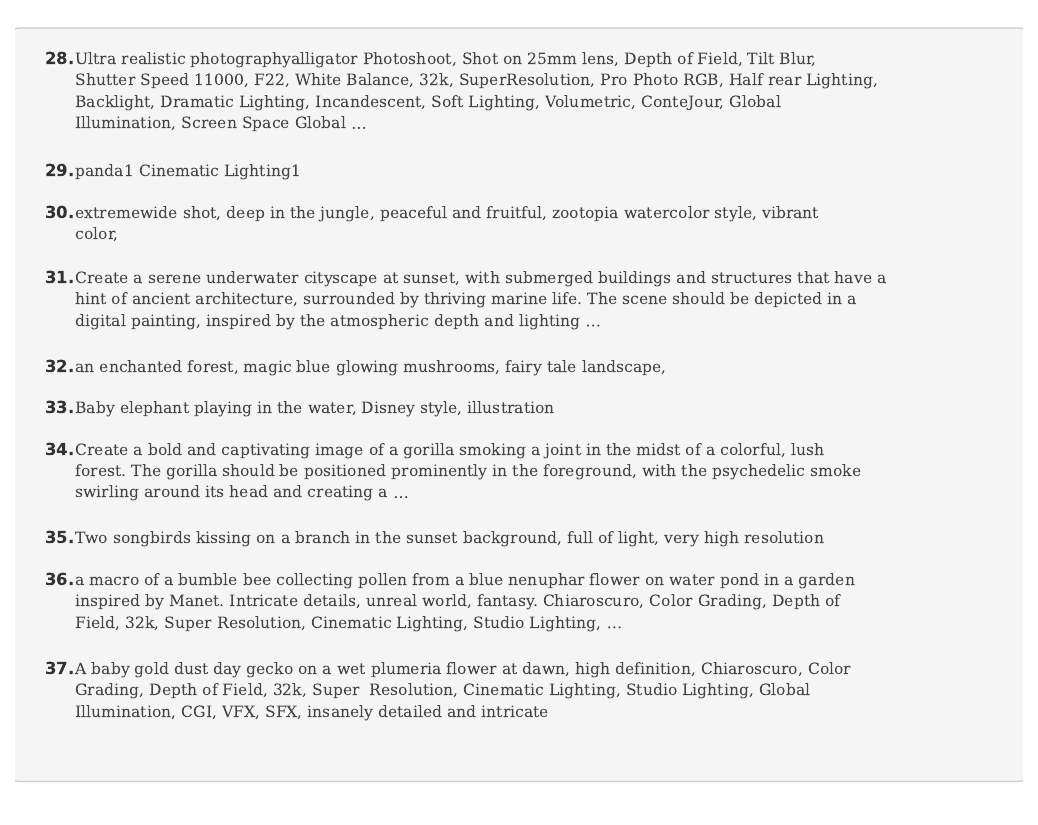}
\vskip -0.05in
\caption{\textbf{Text prompts for T2I qualitative samples (3/3).}}
\label{fig:qualitative_t2i_prompts3}
\end{figure*}

\Paragraph{Text-to-image generation.}
We additionally show text-to-image samples from RAEv2 (0.9B) in \fref{fig:qualitative_t2i}--\fref{fig:qualitative_t2i3}. The model (0.9B) is trained for 100K iterations with batch size 1024 and evaluated on MJHQ test set prompts, generating at 256$\times$256 resolution using self-guidance with the REPA head. Despite the relatively short training schedule and small model size, the samples demonstrate strong prompt adherence across a range of subjects including animals, landscapes, and stylized scenes. The corresponding text prompts are listed in \fref{fig:qualitative_t2i_prompts}--\fref{fig:qualitative_t2i_prompts3}.

\section{Discussion and Limitations}
\label{sec:discussion}

We next provide a discussion of some of the limitations of the current work, which might motivate further research in this area. We only consider very simple approaches for Generalized Representation Autoencoders. In particular, we only consider simple addition and random projection as one of the key ways for aggregating features across different layers of a pretrained vision encoder. In future, better optimization of the aggregation recipe can provide further gains for both generation and reconstruction.

Also similar to iREPA \cite{irepa}, we identify the best representation for RAE through empirical search over a discrete set of pretrained encoders. In future work, we would like to directly optimize the representation itself for better generation, with end-to-end learning \cite{repae}.

\addtocontents{toc}{\protect\setcounter{tocdepth}{-1}}
\section{Note on LLM Usage}
\label{sec:appendix_llm_usage}

All figures in the paper are directly generated from our experiment logs and checkpoints using Claude Code (Anthropic, 2025). Additionally, we use LLM help for searching and formulating relevant work in \sref{sec:appendix_related}. We also use Cursor in some parts to help with paper writing.

\end{document}